# A Comprehensive Survey on Model Quantization for Deep Neural Networks in Image Classification

BABAK ROKH

Department of Computer Engineering, University of Zanjan, Zanjan, Iran, babak.rokh@znu.ac.ir

ALI AZARPEYVAND

Department of Electrical and Computer Engineering, University of Zanjan, Zanjan, Iran, azarpeyvand@znu.ac.ir

ALIREZA KHANTEYMOORI

Neurozentrum Department, Universitätsklinikum Freiburg, Freiburg, Germany, alireza.khanteymoori@uniklinik-freiburg.de

Recent advancements in machine learning achieved by Deep Neural Networks (DNNs) have been significant. While demonstrating high accuracy, DNNs are associated with a huge number of parameters and computations, which leads to high memory usage and energy consumption. As a result, deploying DNNs on devices with constrained hardware resources poses significant challenges. To overcome this, various compression techniques have been widely employed to optimize DNN accelerators. A promising approach is quantization, in which the full-precision values are stored in low bit-width precision. Quantization not only reduces memory requirements but also replaces high-cost operations with low-cost ones. DNN quantization offers flexibility and efficiency in hardware design, making it a widely adopted technique in various methods. Since quantization has been extensively utilized in previous works, there is a need for an integrated report that provides an understanding, analysis, and comparison of different quantization approaches. Consequently, we present a comprehensive survey of quantization concepts and methods, with a focus on image classification. We describe clustering-based quantization methods and explore the use of a scale factor parameter for approximating full-precision values. Moreover, we thoroughly review the training of a quantized DNN, including the use of a straight-through estimator and quantization regularization. We explain the replacement of floating-point operations with low-cost bitwise operations in a quantized DNN and the sensitivity of different layers in quantization. Furthermore, we highlight the evaluation metrics for quantization methods and important benchmarks in the image classification task. We also present the accuracy of the state-of-the-art methods on CIFAR-10 and ImageNet. This paper attempts to make the readers familiar with the basic and advanced concepts of quantization, introduce important works in DNN quantization, and highlight challenges for future research in this field.

## 1 INTRODUCTION

Deep Neural Networks (DNNs) have grown significantly in the recent decade, yielding considerable results in various areas of machine learning, such as image classification [1, 2], Natural Language Processing (NLP) [3-5], and speech recognition

[6, 7]. First, in the 2012 ImageNet challenge, AlexNet [1], a Deep Convolutional Neural Network (DCNN), achieved the highest accuracy among all the models. In subsequent years, DCNNs were further developed and achieved higher accuracy in the ImageNet challenge [1, 2, 8]. By adding more layers, they even outperformed humans in terms of image classification accuracy in some cases [8, 9]. However, DCNNs have a huge number of parameters for storage and heavy computations, posing challenges the use of them on devices with limited hardware resources. Table 1 presents the storage and computational specifications of several popular DCNNs. The main operation in DCNNs is multiply-accumulate (MAC) in convolution and Fully-Connected (FC) layers.

Table 1: Specifications of several DCNNs

| DCNN | Input size | Number of Conv. layers | Number of FC layers | Number of weights | Total MACs ($\times 10^9$) |
| --- | --- | --- | --- | --- | --- |
| AlexNet [1] | 227×227 | 5 | 3 | 61M | 0.72 |
| OverFeat [10] | 231×231 | 5 | 3 | 146M | 2.8 |
| VGGNet-16 [11] | 224×224 | 13 | 3 | 138M | 15.5 |
| GoogleNet (Inception-V1) [2] | 224×224 | 57 | 1 | 7M | 1.43 |
| ResNet-50 [8] | 224×224 | 49 | 1 | 25.5M | 3.9 |
| SqueezeNet [12] | 224×224 | 26 | 0 | 1.2M | 1.7 |
| MobileNetV1 [13] | 224×224 | 27 | 1 | 4.2M | 0.57 |
| ShuffleNet [14] | 224×224 | 49 | 1 | 1.4M | 0.14 |

Accordingly, accelerating DNNs processing is essential. Consequently, the topic "acceleration and compression of DNNs" has widely raised recently. In designing accelerators, researchers concentrate on network compression, parallel processing, and optimizing memory transfers for processing speed-up [15-39].

In the beginning, the focus was on hardware optimization for processing speed-up in DNN accelerators [26, 27, 40, 41]. Later, researchers such as Song Han *et al.* [17], Matthieu Courbariaux *et al.* [21], Mohammad Rastegari *et al.* [22], and Shuchang Zhou *et al.* [23], concluded that compression and software optimization of DNNs can be more effective before touching hardware. Therefore, various approaches have been proposed for DNN compression to eliminate unnecessary parameters and computations with acceptable accuracy [17-19, 21-25, 42-46]. Figure 1 shows the design stages for two generations of DNN accelerators.

Compression techniques have been utilized in neural networks since the end of the 1980s [47, 48]. As the use of DNNs on devices with constrained hardware resources has evolved, compression techniques have emerged as promising approaches to reduce parameters and computations, resulting in a reduction of memory access which consumes high energy. A smaller model is saved on energy-efficient on-chip memory rather than power-hungry off-chip DRAM. Moreover, the deployment of a compressed model on hardware becomes cheaper since it requires fewer hardware resources and a smaller chip die area. Figure 1b represents the approaches in DNN compression, which are listed as

1. **Quantization** approximates the numerical network components with low bit-width precision [17, 21-25, 42]. For instance, 32-bit floating-point weights can be mapped to a 16-bit or 8-bit integer.
2. **Pruning** is one of the first approaches for compressing neural networks. This technique is used for removing unnecessary or less important connections within the network and making a sparse network that reduces memory usage as well as computations. Papers [47, 49, 50] are among the first network pruning works done on shallow neural networks. In recent years, this approach has been widely used in DNN compression [17-20, 44, 45, 51-60].

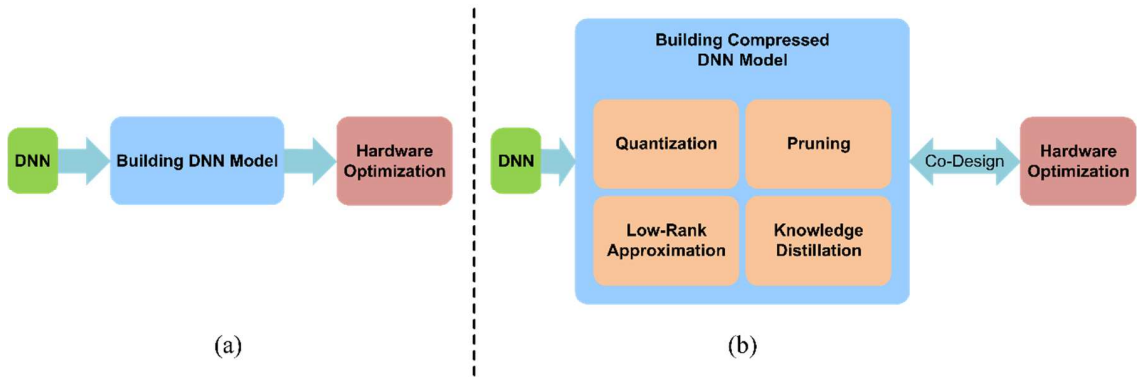

Figure 1: Design stages for two generations of DNNs accelerators a) The stages of designing the first generation of DNNs accelerators b) Co-designing of the compressed model and the hardware optimization in the second generation of DNNs accelerators

3. **Low-rank approximation**, as an approach to simplify matrices and images, creates a new matrix close to the weight matrix, which has lower dimensions and fewer computations in DNNs. Works [46, 61-66] employed the low-rank approximation approach for DNN compression.
4. **Knowledge Distillation (KD)**, also known as teacher-student, which a complex model as a teacher trains a simple model as a student, and finally, the complex model is replaced with the simple one. In this approach, the teacher model is trained, and then the knowledge of the teacher model is used for training the student model. The goal of KD is to employ a simpler model that exhibits generalization and accuracy comparable to the complex model. Thus, the KD approach is effective in reducing the number of parameters and computations in DNNs. Works [34-39] used the KD technique for DNN compression.

Some works employed a combination of multiple compression approaches to achieve greater efficiency. For instance, works [17, 67, 68] used a combination of pruning and quantization, and [69] utilized a combination of KD and quantization. Some works suggested joint optimization of compression approaches instead of their independent optimization to further enhance the effectiveness of combining these compression techniques [70-72]. In this work, we focus on quantization methods for DCNN compression.

One of the important approaches to compress DNNs is quantization, which is frequently employed for reducing memory usage and implementing optimal hardware in various applications. The quantization of neural networks dates back to the beginning of 1990s when some works adopted the quantization of weights in shallow networks [73-76]. Since 2014, numerous methods have been proposed for DNN quantization [21-25]. In this work, we survey quantization methods for DNNs from different perspectives and subsequently mention the advantages and challenges of DNN quantization. Some significant advantages of quantization are:

1. High compression is achieved by quantization compared to other approaches, whereas the accuracy reduction is less [77].
2. One of the reasons for the widespread use of quantization is its flexibility. Since quantization is not dependent on the network architecture, a quantization algorithm can be applied to various types of DNNs. Many quantization methods originally designed for DCNNs are also used for Recurrent Neural Networks (RNNs) and Long Short-Term Memory (LSTM) networks [78-80].

3. In quantization, high-cost floating-point operations are replaced with low-cost operations that require a smaller number of cycles on hardware, such as FPGAs.
4. Quantization reduces the cost of hardware accelerator design. For instance, in 1-bit quantization, a 32-bit floating-point multiplier can be replaced with an XNOR operator, leading to a cost reduction of 200 times in Xilinx FPGA [78]. Paper [81] gives an overview of quantization techniques on FPGA, discussing different strategies and addressing the challenges associated with their implementation.
5. As quantization simplifies parameters for the network, it can contribute to controlling overfitting.

Only a small fraction of recent survey papers have specifically focused on quantization methods. The majority of these surveys have reviewed various approaches to DNN compression and acceleration, allocating only a small section to the topic of quantization [82-95]. Paper [96] focused on the pruning and quantization approaches and discussed some basic concepts of them. Papers [97-100] surveyed quantization methods. Qin *et al.* [97] almost have a comprehensive discussion on binary quantization. Rakka *et al.* [98] concentrated on mixed-precision quantization and presented a comprehensive review of this approach's methods. Guo *et al.* [99] reviewed quantization methods from four perspectives: 1) network components that can be quantized, 2) deterministic and stochastic quantization, 3) fixed and adaptive codebook quantization, and 4) quantization during and after training. Gholami *et al.* [100] explained uniform and non-uniform quantization, distillation-assisted quantization, hardware-aware quantization, and extreme quantization (quantization in the low bit-width) and referred to quantization on processors.

The present paper concentrates on the methods of DCNN quantization for the image classification task, mentioning the concepts of quantization thoroughly and reviewing the previous methods from different perspectives in various categories. We describe using the scale factor and the clustering-based approaches for approximating the quantization levels accurately. Additionally, the training of a quantized neural network and using a Straight-Through Estimator (STE) in the Error Back Propagation (EBP) algorithm is reviewed comprehensively. The effect of quantization on the MAC operation of DCNNs and the sensitivity of DCNNs layers in quantization, and mixed-precision quantization are discussed. Finally, we denote the common benchmark datasets for evaluating the methods and present the results of state-of-the-art (SOA) methods.

The organization of the remaining sections is summarized as follows. In section 2, preliminary concepts of neural networks are described. In section 3, the concepts of quantization are explained. Section 4 details the training of a quantized network. Section 5 describes the operations in a quantized DCNN. In section 6, the sensitivity of DCNN layers to quantization is discussed. Section 7 presents the evaluation of SOA methods and an analysis of the results. Finally, section 8 offers conclusions and suggestions for future research.

## 2 PRELIMINARY CONCEPTS OF NEURAL NETWORKS

A neural network is an acyclic graph consisting of nodes (neurons) organized in multiple layers. Each layer connects to another layer through neurons. In general, each neural network has three types of layers: 1) an input layer, 2) multiple hidden layers, and 3) an output layer. Figure 2 shows a typical neural network. Neurons in hidden layers include an activation function, while neurons in the output layer do not have an activation function or are thought to have a linear activation function.

### 2.1 Convolutional Neural Networks

Depending on the application, there exist various types of neural networks with different architectures. One of the most widely used neural networks is the Convolutional Neural Network (CNN), which is inspired by the visual cortex organization in humans. The first CNN, Time delay Neural Network (TNN), was introduced by Alexander Waibel in 1989 for speech recognition [101]. Then, in 1998, Yann LeCun *et al.* presented LeNet-5, which is regarded as the turning point in the development of CNNs [102]. CNNs extract high-level and more complex features by combining low-level and simple features [103]. This process is similar to how the humans visual system performs. As CNNs were originally designed for image processing, in their architecture is assumed that the input is an image [104].

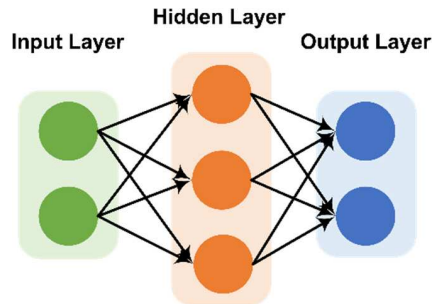

Figure 2: A typical neural network

DCNNs have demonstrated superior accuracy, but they also have a significant number of computations. For this reason, Graphics Processing Units (GPUs) have been applied for processing these networks. Advances in GPUs have led to the development of DCNNs and the possibility of increasing the number of layers. Following that, considerable efforts were devoted to the development of DCNNs, and since 2012, all the winning models in the ImageNet challenge have been DCNNs [105].

A DCNN consists of various types of layers, and the common layers include convolution layer, normalization layer, pooling layer, and FC layer. Most parameters and computations are in FC and convolution layers. The main layer in DCNN is the convolution layer, which is formed in three dimensions. This layer produces an output feature map by convolving multiple filters (weights) with the input feature map. There is weight sharing in the convolution layer, which means that each weight is applied to different connections. The majority of computations in DCNNs are in this layer due to its three-dimensional structure and weight sharing. Figure 3 illustrates an example of the convolution process, where an input feature map with the size of $C_{in} \times W_{in} \times H_{in}$ is convolved with filters of size $C_{in} \times K_h \times K_w$, resulting in an output feature map with the size of $C_{out} \times W_{out} \times H_{out}$. Here, $C_{in}$ and $C_{out}$ represent the size of the input channel and the size of the output channel in the layer, respectively. The size of the channel in the input feature map and the filters is the same, and the number of filters in a layer determines the size of the output channel for that layer. Weight sharing in the convolution layer leads to a significant reduction in the number of parameters. In contrast, the majority of parameters in DCNNs are typically in the FC layers, where each neuron is connected to all neurons in both the previous and next layers. As the convolution and FC layers contain the majority of computations and parameters in DCNNs, the primary focus is on these layers in accelerators and compression techniques.

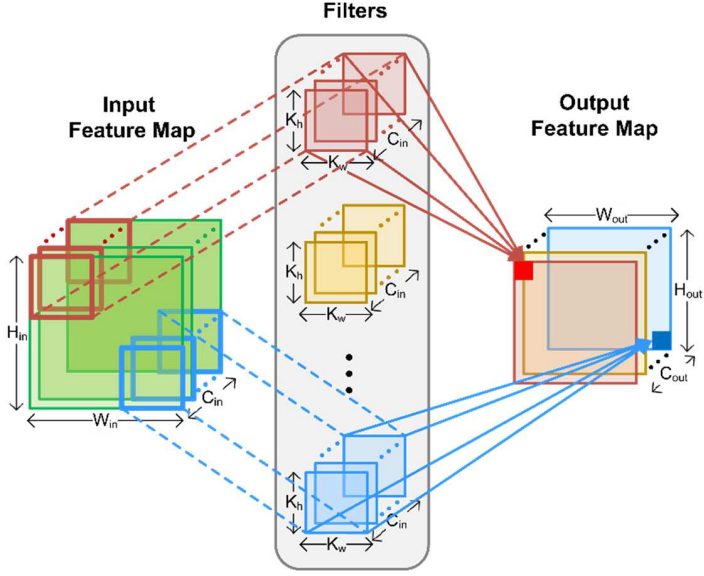

Figure 3: Three-dimensional structure of the convolution layer, and convolution of filters with patches of the input feature map

## 3 CONCEPTS OF QUANTIZATION

Quantization is mapping values from a continuous space to a discrete space, where full-precision values are mapped to new values with lower bit-width called quantization levels. This mapping is performed using a constant piecewise function, such as

$$Q(r) = q_i \quad r \in [r_i, r_{i+1}), \ i = 1, \ldots, m \tag{1}$$

Accurate quantization relies on the computation of optimal quantization levels. Figure 4 shows a quantization, mapping continuous full-precision values in the range $[r_1, r_5]$ to a limited set of discrete quantization levels $Q = \{q_1, q_2, q_3, q_4\}$. The quantization function $Q(r)$ maps the range $[r_1, r_2]$ to quantization level $q_1$, the range $[r_2, r_3]$ to $q_2$, the range $[r_3, r_4]$ to $q_3$, and the range $[r_4, r_5]$ to $q_4$. The step size specifies the distance between two successive quantization levels as Equation (2). In the following, various types of quantization for neural networks are discussed from different perspectives.

$$step\ size = q_{i+1} - q_i \tag{2}$$

### 3.1 Network Components and Quantization

Each numerical component in neural networks can be quantized. These components are typically divided into three main categories: weights, activations, and gradients. Each parameter in the neural networks can be quantized, and quantization of the weights is the most common. In most cases, biases and other parameters, such as batch normalization parameters, are kept in full precision in view of the fact that they include a minimal rate of neural network parameters, and the quantization of them is less efficient in compression. Nevertheless, some works have performed the quantization of these parameters, such as paper [106] which represents biases with 1-bit precision.

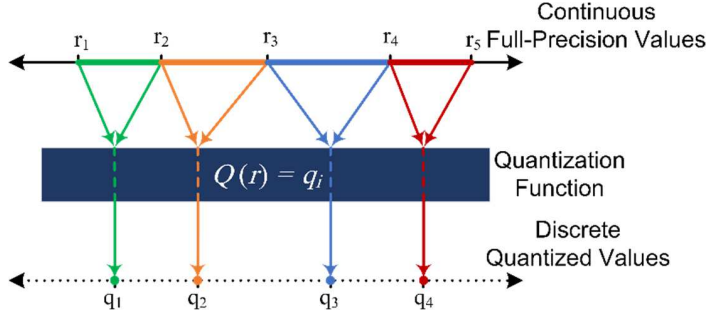

Figure 4: Mapping continuous full-precision values to discrete quantized values using a quantization function

Activations quantization poses additional challenges compared to weights quantization. While weights remain fixed after training, activations change during the inference phase according to the input data. At first, most methods focused only on weight quantization. However, weight quantization alone while keeping the activations in full precision limits the quantization efficiency, especially in the CNNs, where the values of weights in the convolution layers are relatively smaller compared to the FC layers [107]. Moreover, weight sharing in convolution layers reduces the number of weights in these layers, resulting in lower memory usage than activations [108]. In recent years, most methods have suggested the quantization of both weights and activations. Storing both weights and activations in low bit-width enables MACs to be performed using low-cost operations, resulting in reduced computational cost and memory usage.

Gradients quantization is merely efficient for speed-up in the training phase. The quantization of gradients is more challenging than weights and activations quantization. Gradients are propagated from the output to the first layer of a network during the backward pass of the EBP algorithm. High-precision gradients are essential for the convergence of the optimization algorithm during training. Furthermore, due to the wide range of gradient values, accurate quantization requires the use of more bits [109].

Almost all works that have performed activations quantization and gradients quantization have also included weight quantization in their methodologies. In one experiment conducted in work [110], activations were quantized while the weights were kept in full precision. Table 2 presents information about the quantized components in several previous works [17, 21-25, 42, 71, 78- 80, 106, 107, 109-127].

### 3.2 Time of Quantization

There are two general approaches for quantization based on the timing of when quantization is performed: Quantization-Aware Training (QAT) and Post-Training Quantization (PTQ). QAT is conducted during training, whereas PTQ is performed after training.

*3.2.1 Quantization-Aware Training*

QAT is in such a way that the quantization and training are performed simultaneously, and the network is trained with quantized values. In the QAT approach, the network is trained with discrete quantized values. In a low bit-width precision quantized network, the convergence of the learning algorithm is challenging. The training of a quantized network commonly requires more iterations than the full-precision network for convergence. It needs customized solutions compatible with a discrete network. The QAT approach is employed for compression and speed-up in both the training and inference phases.

Table 2: Quantized components in some methods: Weights (W), Activations (Act), and Gradients G) and their type of quantization: uniform (U) or non-uniform (NU), and time of quantization: QAT or PTQ

| Method | Components Quantization | | | Time of Quantization |
|---|---|---|---|---|
| BinaryConnect [21] | W (U) | | | QAT |
| TernaryConnect [109] | W (U) | | | QAT |
| Sung *et al.* [111] | W (U) | | | PTQ |
| DeepCompression [17] | W (NU) | | | PTQ |
| Bitwise Neural Networks [106] | W (NU) | Act (NU) | | QAT |
| TWN [112] | W | | | QAT |
| BNN [42] | W (U) | Act (U) | | QAT |
| XNOR-Net [22] | W | Act | G | QAT |
| Dorefa-net [23] | W | Act | G | QAT |
| TTQ [113] | W | | | QAT |
| Miyashita *et al.* [110] | W (NU) | Act (NU) | G(UN) | QAT |
| QNN [78] | W (U) | Act (U) | G(U) | QAT |
| LogNet [114] | W (NU) | Act (NU) | | PTQ |
| Tang *et al.* [115] | W (U) | Act | | QAT |
| HWGQ [116] | W | Act | | QAT |
| ABC [117] | W | Act | | QAT |
| INQ [118] | W (NU) | | | PTQ |
| FGQ [119] | W (U) | Act | | PTQ |
| WQ [80] | W (NU) | Act (NU) | | QAT |
| Balanced quantization [79] | W (NU) | Act (NU) | G(F/NU) | QAT |
| Zhuang *et al.* [24] | W (NU) | Act (NU) | | QAT |
| PACT [107] | W | Act | | QAT |
| Xu *et al.* [120] | W (NU) | | | PTQ |
| ELQ [121] | W | | | PTQ |
| LQ-Nets [122] | W (U) | Act (U) | | QAT |
| BitPruning [123] | W (U) | Act (U) | | QAT |
| APoT [124] | W (NU) | Act (NU) | | QAT |
| Bi-RealNet [25] | W | Act (U) | | QAT |
| IR-Net [125] | W (U) | Act (U) | | QAT |
| DJPQ [71] | W (NU) | Act (U) | | QAT |
| DMBQ [126] | W (U) | Act (U) | | QAT |
| SQ [127] | W(U) | | | QAT |

### 3.2.2 Post-Training Quantization

PTQ is performed after network training and building the model with full-precision values. Therefore, this approach is employed for compression and speed-up in the inference phase. Weight quantization through PTQ often leads to a reduction in model accuracy. To address this issue, retraining is performed after quantization to improve accuracy, and hence, the PTQ methods have proposed fine-tuning after quantization for compatibility of the trained model with the quantization process. After quantization, the network is retrained with the quantized weights once or repeatedly to reach an acceptable accuracy. Sung *et al.* [111] showed that there is a big gap between the model accuracy in PTQ quantization with and without retraining. Figure 5 demonstrates the overall steps in the PTQ approach.

The model accuracy in the QAT approach is commonly higher than in PTQ because the trained model is more compatible with the quantization process. Table 2 provides information on the time of quantization in some previous works.

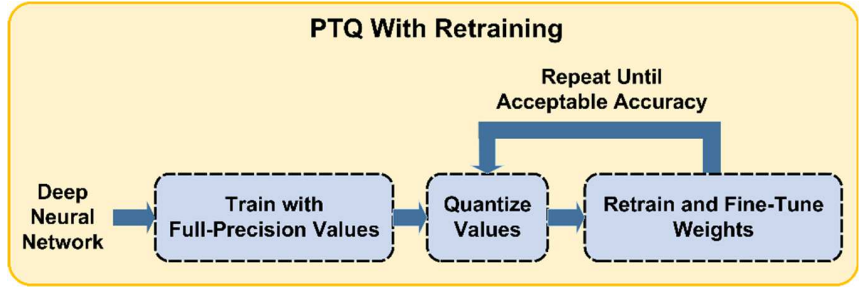

Figure 5: PTQ with retraining until reaching acceptable accuracy

## 3.3 Deterministic and Stochastic Quantization

Full-precision values can be mapped to quantization levels either deterministically or stochastically. In deterministic quantization, each full-precision value is consistently mapped to a specific quantization level. On the other hand, in stochastic quantization, a full-precision value may be mapped to each of the quantization levels with a probability, although it is more likely to encode a specific quantization level that has the highest probability.

*3.3.1 Deterministic Quantization Methods*

One of the deterministic functions commonly used for binary quantization is the *Sign* function:

$$b = Sign(x) = \begin{cases} +1 & x \geq 0 \\ -1 & x < 0 \end{cases} \qquad (3)$$

In Equation (3), the full-precision values ($x$) are mapped to the binary values: +1 or -1. Several papers, such as [22, 25, 42, 100, 109, 110], have employed Equation (3) for quantization. Deterministic ternary quantization is defined as

$$t = \begin{cases} +1 & x > \Delta \\ 0 & |x| \leq \Delta \\ -1 & x < -\Delta \end{cases} \qquad (4)$$

In Equation (4), $\Delta$ represents a threshold. Papers [112, 119, 121, 128] have utilized Equation (4) for quantization. Paper [111] proposed a deterministic quantization as Equation (5).

$$Q(x) = Sign(x).d.\min\left(round(\tfrac{|x|}{d}), \tfrac{M-1}{2}\right) \qquad (5)$$

In Equation (5), $d$ represents the step size, and $M$ is an odd number and determines the number of quantization levels. Consequently, the quantization levels include zero, positive and negative values symmetrically. Papers [23, 24, 79] employed a deterministic quantization with $k$ bit-width. First, they mapped the values to the range [0,1], then used Equation (6) for quantization with $k$ bit-width.

$$quantize_k = \tfrac{1}{2^k-1} round\left((2^k-1)x\right), \quad 0 \leq x \leq 1 \qquad (6)$$

Equation (6) maps the full-precision values in the range $x \in [0,1]$ to $2^k$ quantization levels within the same interval with step size $\tfrac{1}{2^k-1}$. For $k$ bit-width, the quantization levels are $L_q = \left\{0, \tfrac{1}{2^k-1}, \tfrac{2}{2^k-1}, \ldots, 1\right\}$. For example, for $k$=2, there are $2^2$=4 quantization levels which are $L_q$ = {0, 1/3, 2/3, 1}.

In the Learned Quantization Network (LQ-Nets) method [122], a deterministic quantization is used, in which both weights and activations are quantized in $k$ bits, and each bit can take the values of 1 or -1. Consequently, there are $2^k$ quantization levels ($q_1<q_2<...<q_{2^k}$). Each quantization level is obtained as the result of an inner product between a basis vector and a $k$-bit binary vector:

$$Q(x,v) = v^T e_l \quad e_l \in \{-1,1\}^K, \; x \in (t_l, t_{l+1}) \tag{7}$$

In Equation (7), $x$ represents full-precision values, $v \in \mathbb{R}^K$ denotes the learnable floating-point basis vector, and $e_l$ is a $k$-bit binary vector from [-1, -1, ..., -1] to [1, 1, ..., 1].

*3.3.2 Stochastic Quantization Methods*

Few previous research works used stochastic quantization [21, 78, 129]. The BinaryConnect method [21] used a stochastic binary quantization:

$$b = \begin{cases} +1 & p = \sigma(x) \\ -1 & q = 1 - p \end{cases} \tag{8}$$

In Equation (8), $p$ and $q$ determine the probabilities of mapping to 1 or -1, respectively. $\sigma$ is the *hard sigmoid* function as

$$\sigma(x) = clip\left(\frac{x+1}{2}, 0, 1\right) = \max\left(0, \min\left(1, \frac{x+1}{2}\right)\right) \tag{9}$$

Equation (9) assigns a probability to the full-precision values in the range (-1,1). For $x \in (0,1)$, the probability is bigger than 0.5, while it is smaller than 0.5 for $x \in (-1, 0)$. When $x$ is equal to zero, the probability is 0.5. Therefore, according to Equation (8), the full-precision values belonging to the interval (0,1) are more likely to be mapped to +1, with a higher probability for values closer to 1. Conversely, for the interval (-1,0), there is a higher probability of mapping to -1, with the probability increasing for values closer to -1.

Lin *et al.* changed Equation (8) for stochastic ternary quantization with {-1, 0, 1} quantization levels in the TernaryConnect method [109]. The values in the interval [-1,1] are mapped to a quantization level by the following probabilities:

$$\begin{cases} \text{if } w > 0: & p(t = 1) = w; \quad p(t = 0) = 1 - w \\ \text{if } w < 0: & p(t = -1) = -w; \quad p(t = 0) = 1 + w \end{cases} \tag{10}$$

In Equation (10), $w$ and $t$ represent full-precision and ternary weights, respectively.

*3.3.3 Deterministic and Stochastic Quantization Comparison*

Stochastic quantization has shown better model generalization compared to deterministic quantization [21, 129]. In stochastic quantization, a parameter is not mapped to a definite quantization level, which means that relying on specific features is not possible. Instead, the weights are spread out like a regularizer, thereby promoting generalization. Implementation of stochastic quantization is more challenging and costly than deterministic quantization, particularly in hardware implementations, as it requires a random bit generator [42].

**3.4 Quantization Levels Based on Distribution**

An efficient quantization represents the distribution of full-precision values and preserves informative parts of the original data. In quantization, multiple values are limited to one value, and determining quantization levels is crucial to minimize quantization accuracy degradation. In this section, the methods for determining quantization levels based on the distribution of full-precision values are discussed.

*3.4.1 Uniform and Non-uniform Quantization*

By adjusting the step size, the distribution of quantization levels is changed, and the quantization is divided into two general categories: uniform and non-uniform. In uniform quantization, the step size is constant, while in non-uniform quantization, it varies. Figures 6a and 6b show uniform and non-uniform quantization, respectively. All the quantization functions introduced in section 3.3 are uniform. Uniform quantization is generally simpler to implement compared to non-uniform quantization. In non-uniform quantization, the step size is determined according to the distribution of the full-precision values, which makes it more complex and accurate than uniform quantization.

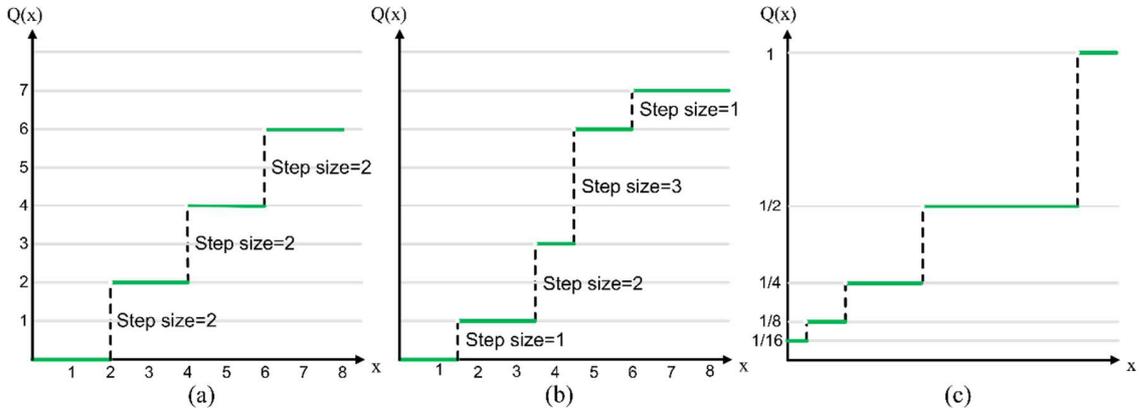

Figure 6: Quantization types based on step size adjustment a) Uniform quantization with the same step sizes b) Non-uniform quantization with different step sizes c) Base-2 logarithmic quantization

Logarithmic quantization is a type of non-uniform quantization with exponential quantization levels. Figure 6c shows a base-2 logarithm quantization, and Equation (11) presents its relation.

$$Q(x) = Sign(x) 2^{round(\log_2 |x|)} \tag{11}$$

In Equation (11), $x$ represents full-precision values. Logarithmic quantization allows the encoding of a larger range of numbers using the same storage in comparison with uniform quantization by storing a small integer exponent instead of a floating-point number.

Previous studies have revealed that weights in DCNNs often follow a normal distribution with a mean of zero [80]. For instance, Figure 7 illustrates the distribution of weights in the convolution layers of the trained MobileNetV2 on ImageNet, showing a normal distribution in the range [-0.4, 0.4] with a mean close to zero. In logarithmic quantization, the quantization levels are denser for values close to zero. Therefore, the distribution of quantization levels in logarithmic quantization is matched to the distribution of the full-precision weights in DCNNs, which leads to more accurate quantization.

The base-2 logarithm quantization is naturally a representation of the binary system. As a result, it is well-matched to digital hardware and provides simple operations. Works [110, 114, 118, 123, 130] employed logarithmic quantization. Table 2 summarizes the previous methods that have applied uniform or non-uniform quantization approaches to weights, activations, and gradients separately.

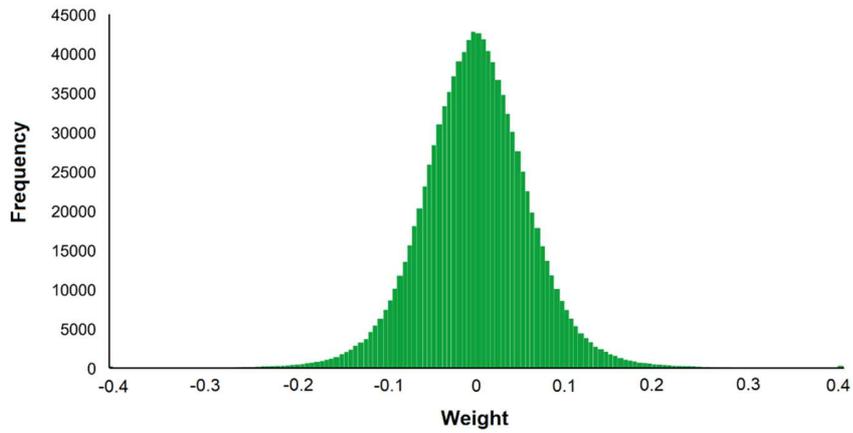

Figure 7: Distribution of the weights in convolution layers of the trained MobileNetV2 on ImageNet

*3.4.2 Clustering-Based Quantization*

Quantization and clustering share similarities. In clustering, each value is assigned to a cluster, while in quantization, each full-precision value is mapped to one of the quantization levels. Many works have used clustering techniques for weight quantization. In clustering-based methods, weights that belong to the same cluster are represented by a single value. The number of quantization levels is equal to the number of clusters. After clustering, instead of storing the floating-point weights (center of the clusters), the cluster index is stored as an integer number, resulting in reduced memory requirements. Figure 8 shows an example of 2-bit clustering-based quantization, where the weights are grouped into four clusters, and the cluster index can be represented using 2 bits in the codebook.

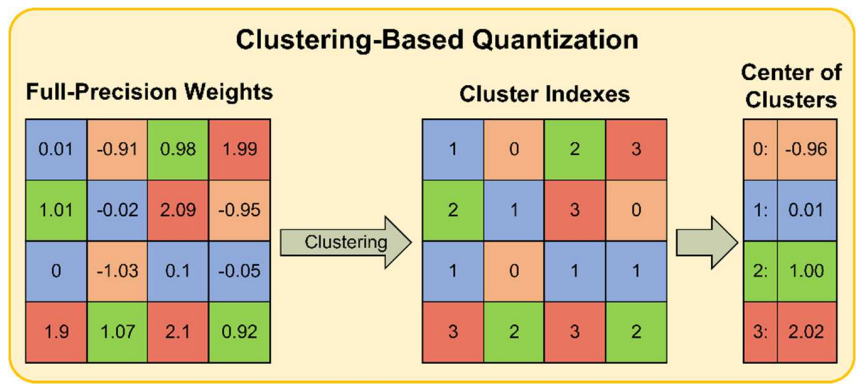

Figure 8: Weights grouping into four clusters

In the DeepCompression method [17], the weights are clustered using the k-means algorithm, where the weight values in a cluster are close to each other and mapped to the same quantization level, which is the cluster center. Xu *et al.* proposed two clustering-based methods for high and low bit-width quantization. The first method, called Single Level Quantization (SLQ), is used for weight quantization with high bit-width precision [120]. In SLQ, the weights of each layer are clustered separately using the k-means algorithm. After clustering, the clusters are grouped into two categories based on quantization loss. The clusters with small quantization loss (category 1) are quantized, where all the weights in a cluster are mapped to the center of the related cluster. The clusters in the second category are retrained. These steps are repeated until all the weights are quantized. SLQ is not suitable for low bit-width quantization due to the small number of clusters, which leads to significant quantization loss. Therefore, they suggested Multiple Level Quantization (MLQ) [120] for 2-bit and 3-bit quantization.

The SLQ method partitions weights in a single layer, in other terms, in the width of the network. In contrast, MLQ partitions weights not only in the width but also in the depth of the network. In the depth partitioning, the weights in each layer are clustered, and then, the clusters, called boundaries, with a higher impact on model accuracy, are quantized. Unlike the SLQ method, in which all boundaries in all layers are quantized at the same time, they are quantized iteratively and incrementally in MLQ. The idea behind this incremental approach comes from the different effects of layers on the accuracy of a quantized model. The clusters in each boundary (clusters in width) are quantized using the same clustering technique as in SLQ, which involves grouping the clusters into two categories. The iterations continue until all clusters in the depth are quantized. Xu *et al.* also suggested Extended Single Level Quantization (ESLQ) [120], which changes cluster centers as quantization levels to values with a specific type. For example, quantization levels are mapped to the closest number in the form of Power Of Two (POT), making it well-suited for implementation on FPGA platforms.

Park *et al.* [80] suggested allocating more clusters to the informative parts to preserve the information after quantization. They introduced a weighted entropy measure for evaluating the quality of clustering, which determines the importance of full-precision values on network performance. For *N* clusters, weighted entropy is defined as

$$S = -\sum_n I_n P_n \log P_n \tag{12}$$

where

$$P_n = \frac{|C_n|}{\sum_k |C_k|} \tag{13}$$

$$I_n = \frac{\sum_m i_{(n,m)}}{|C_n|} \tag{14}$$

In Equations (13) and (14), Cn denotes the number of weights in cluster n. Equation (13) indicates the density of cluster n. In Equation (14), $i_{(n,m)}$ is defined as Equation (15), where w(n,m) refers to weight m in cluster n.

$$i_{(n,m)} = (w_{(n,m)})^2 \tag{15}$$

Equation (15) determines the importance of the weights, considering that larger weights have a more significant impact on the output. Equations (13) to (15) show that the clustering is performed based on the density of the weights. The result of Equation (12) is maximized by finding the optimum weights clustering. However, this process is very time-consuming due to the need to test various states of weights clustering.

**Challenges of the clustering-based approach**: While clustering algorithms are effective at mapping optimal groups of weights to quantization levels, they are not suitable for implementation in hardware and software due to their significant time complexity and computational requirements for codebook reconstruction [131]. Moreover, in a clustering-based approach, the weights within a cluster are not contiguous in memory, which leads to irregular memory accesses with long delays [119].

The clustering-based approach is primarily performed for weight quantization. Weights remain fixed after network training and are used during inference, while activations change during inference; thus, the clustering-based approach is not suitable for activations quantization.

*3.4.3   Scale Factor*

Using a scale factor parameter is common for accurately approximating the quantization levels. It is especially effective in low bit-width quantization where the number of quantization levels is limited. The scale factor parameter is multiplied by the quantization levels and is determined in such a way that the mathematical expectation of the full-precision values is close to the mathematical expectation of the related quantization levels. This helps in mapping the quantization levels to dense and informative regions of the full-precision values. The scale factor is typically computed based on the mean of the full-precision values.

The Ternary Weight Network (TWN) method [112] computed a scale factor and a threshold for each layer to approximate accurate ternary weights. The optimum scale factor and threshold were computed to minimize the difference between the quantized weights and the corresponding full-precision weights.

$$\alpha^*, \Delta^* = \min_{\alpha \geq 0, \Delta > 0} ||W - \alpha W^t||_2^2 \tag{16}$$

In Equation (16), $\alpha$ and $\Delta$ represent the scale factor and threshold, respectively. $W$ and $W^t$ denote the full-precision weights and ternary weights, respectively. In the TWN method, it is proved that the optimum $\alpha$ and $\Delta$ are computed as

$$\alpha^*_\Delta = \frac{\sum_{i \in I_\Delta} |w_i|}{|I_\Delta|}, \quad \Delta^* = \max_{\Delta > 0} \frac{(\sum_{i \in I_\Delta} |w_i|)^2}{|I_\Delta|} \tag{17}$$

In Equation (17), $I_\Delta = \{1 \leq i \leq n : |w_i| > \Delta\}$ and $|I_\Delta|$ is the number of elements in $I_\Delta$. Since the value of $\Delta^*$ is almost the same for different distributions, $\Delta^*$ is computed for all distributions as

$$\Delta^* \approx 0.7 \frac{\sum_{i=1}^n |w_i|}{n} \tag{18}$$

Finally, the ternary weights are computed by:

$$w_i^t = f_i(w_i|\Delta) = \begin{cases} +1 \times \alpha & w_i > \Delta \\ 0 & |w_i| \leq \Delta \\ -1 \times \alpha & w_i < -\Delta \end{cases} \tag{19}$$

In the XNOR-Net method [22], a scale factor is used for the binarization of weights, activation, and gradients. The scale factor for weights and activations is computed as the filter-wise mean of values, while for gradients, the scale factor is computed as

$$g^* = \max(|g^{in}|) \tag{20}$$

In Equation (20), $g^{in}$ represents the gradients from the next layer in the backward pass. Since the range of gradients is wide, Equation (20) preserves maximum variations in all dimensions [22].

In the Dorefa-net method [23], the scale factor for binary weights is computed the same as the XNOR-Net method, but layer-wise instead of filter-wise. Lin *et al.* [117] proposed a linear combination of several binary weights for quantization in the Accurate Binary Convolutional (ABC) method:

$$W \approx \alpha_1 B_1 + \cdots \alpha_M B_M, \quad B_i \in \{-1,1\}^{w \times h \times c_{in} \times c_{out}} \quad (21)$$

In Equation (21), $W$ indicates the weight matrix of a layer with a size of $w \times h \times c_{in} \times c_{out}$, where $w$ and $h$ are the width and length of the filters, and $c_{in}$ and $c_{out}$ are the numbers of input and output channels. $B$ and $\alpha$ specify the binary weights matrix and scale factor, respectively. To find the optimum scale factors in Equation (21), the solution in the TWN method [112] is employed. For activation quantization, a linear combination similar to Equation (21) is utilized, with the distinction that the scale factor in activations quantization is trainable.

**Trained scale factor**: In the Trained Ternary Quantization (TTQ) method [113], the scale factor is computed separately for the positive and negative ternary weights in each layer to efficiently cover the distribution of weights:

$$w_l^t = \begin{cases} +1 \times W_l^p & w_l > \Delta_l \\ 0 & |w_l| \leq 0 \\ -1 \times W_l^n & w_l < -\Delta_l \end{cases} \quad (22)$$

In Equation (22), $W_l$ and $\Delta_l$ represent the full-precision weights and the threshold, respectively. $W_l^p$ and $W_l^n$ are the scale factors for the positive and negative weights, respectively. The scale factor is updated during the training algorithm to achieve the optimum value. In the backward pass of the EBP algorithm, in addition to the gradient used for updating the network weights, an additional gradient is computed for updating the scale factors $W_l^p$ and $W_l^n$. Equations (23) and (24) compute the gradients for updating the weights and scale factors, respectively.

$$\frac{\partial L}{\partial w_l} = \begin{cases} W_l^p \times \frac{\partial L}{\partial w_l^t} & w_l > \Delta_l \\ 1 \times \frac{\partial L}{\partial w_l^t} & |w_l| \leq \Delta_l \\ W_l^n \times \frac{\partial L}{\partial w_l^t} & w_l < -\Delta_l \end{cases} \quad (23)$$

$$\frac{\partial L}{\partial W_l^p} = \sum_{i \in I_l^p} \frac{\partial L}{\partial w_l^t(i)}, \quad \frac{\partial L}{\partial W_l^n} = \sum_{i \in I_l^n} \frac{\partial L}{\partial w_l^t(i)}, \quad I_l^p \{i | w_l(i) > \Delta_i\}, I_l^n = \{i | w_l(i) < -\Delta_i\} \quad (24)$$

Equation (23) shows the effect of the scale factor on the weights update. Furthermore, it is concluded from Equation (24) that the scale factor is updated with respect to the changes in the weights. Two heuristic methods were proposed for determining the layer-wise threshold. The first method computes the threshold according to the maximum absolute value of the weights as

$$\Delta_l = t \times \max(|\widetilde{w}|) \quad (25)$$

In Equation (25), finding the optimal $t$ can be challenging. The second method determines the threshold based on the rate of sparsity in ternary quantization.

**Scale factor for increasing representational capability**: Mellempudi *et al.*, for ternary weight quantization, divided the weights into $k$ subsets and used a separate scale factor for each subset [119]. Furthermore, they employed two distinct thresholds for the positive and negative weights in each subset. The idea behind different thresholds comes from the different distributions of the positive and negative weights, which are not always symmetric around zero. Thus, they used two different thresholds, $\Delta_p$ and $\Delta_n$, in Equation (19). This approach increases the quantization representational capability from 3 levels {-α, 0, +α} to 2k+1 levels. The optimum scale factor and threshold are calculated similarly to the TWN method [112] for each subset separately.

**Maximum weight in computing the scale factor**: In the Explicit Loss-error-aware Quantization (ELQ) method [121], Equation (26) was proposed for computing the scale factor in binary and ternary quantization, considering the maximum weight in the computation:

$$\alpha_l = mean(W_l) + \beta \max(W_l) \tag{26}$$

In Equation (26), $\beta$ is a positive coefficient, which is considered 0.05. The scale factor is computed layer-wise, and $\beta$ controls the influence of maximum weight. To compute the threshold for ternary weights, the scale factor is utilized as

$$t_l = \begin{cases} \alpha_l & W_l > 0.5\alpha_l \\ -\alpha_l & W_l < -0.5\alpha_l \\ 0 & o.w \end{cases} \tag{27}$$

**The effect of scale factor model on convergence**: Liu *et al.*, in the Bi-RealNet method [25], demonstrated the effectiveness of scale factor on the convergence of a quantized model. They performed weight binarization using the *Sign* function. Figure 9a illustrates the distributions of full-precision and binary weights without using the scale factor in ResNet [25]. As seen, the magnitude of the full-precision weights is equal to 0.1, whereas it is 1 for the binarized weights. This significant gap between the full-precision weights and the binarized weights can lead to a problem in achieving convergence during the training of a quantized network.

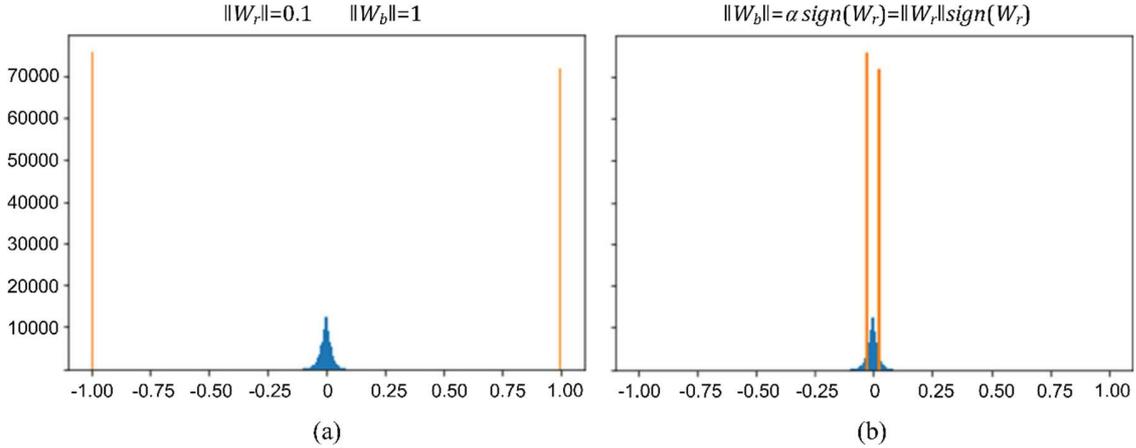

Figure 9: Distribution of the full-precision and binary weights in 18-layer Bi-RealNet. The lines indicate the frequency of binarized weights, a) Without using the scale factor b) Using the scale factor equal to ‖Wr‖ [25]

Liu *et al.* proved that when a BatchNorm layer follows the convolution layer in the network, multiplying all elements in the kernels by a factor of $m$ decreases the gradient by a rate of $1/m$. Consequently, when $m$ is extremely large, the gradient deviates significantly from its true value in the backward pass, which leads to a problem in achieving convergence during training. As observed in Figure 9a, without using a scale factor, the magnitude of the weights increases by a factor of 10 after quantization. The scale factor is used to decrease the distance between the full-precision and quantized weights. In Bi-RealNet, the scale factor is set equal to the magnitude of the full-precision weights. Figure 9b shows that after using the scale factor, the distribution of the full-precision weights becomes close to the distribution of the binarized weights [25].

In the Bi-RealNet method, the scale factor is only computed during network training and not during the inference phase. This is based on the finding that the BatchNorm output is unaffected by the scale factor, and since there are no gradients during inference, there is no need for re-scaling.

**Challenges of using scale factor**: Using a scale factor poses two challenges. First, it requires complex computations. To address this problem, Tang *et al.* [115] proposed an alternative approach where the scale factor is not employed in the quantization function. Instead, they moved the scale factor to the activation function, specifically using *Parametric ReLU* [132] as the activation function:

$$PReLU(x_i) = \begin{cases} x_i & x_i > 0 \\ \alpha_i x_i & x_i \leq 0 \end{cases} \tag{28}$$

In Equation (28), $i$ denotes a channel, and thus, a different $\alpha$ is computed for each channel. The $\alpha$ parameter is updated during training to find the optimum value.

The second challenge of using a scale factor arises from its floating-point value. Using floating-point operations in MACs leads to a large number of cycles on hardware and high computational cost. To overcome this challenge, paper [125] suggested the use of POT numbers as the scale factor. As a result, the floating-point multiplication is replaced with a low-cost bit-shift operator.

## 4 TRAINING OF QUANTIZED NEURAL NETWORK

The main algorithm for training a neural network is EBP [133]. EBP applies the gradient descent algorithm and the chain rule to adjust the network parameters. Suppose that Figure 10 shows an extract of a neural network. In Figure 10, $j$ determines a hidden layer, and the weights between layers $k$ and $j$ are updated as

$$w_{kj}(n) = w_{kj}(n-1) - \gamma \Delta w_{kj}, \quad \Delta w_{kj} = \frac{\partial E}{\partial w_{kj}} = \sum_{i \in I_j}(\delta_i w_{ji}) h'(x_i) y_k = \delta_j y_k \tag{29}$$

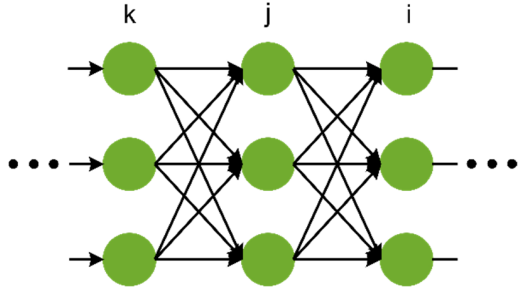

Figure 10: An extract of a neural network with hidden layers i, j, and k

In Equation (29), $w_{kj}(n\text{-}1)$ and $w_{kj}(n)$ indicate the weights between $k$ and $j$ layers before and after the update, respectively. $\gamma$ and $\delta$ are the learning rate and the error signal, respectively. $x_i$ and $y_i$ are the inputs and outputs of layer $i$, respectively, and $h'$ denotes the derivative of the activation function. As this equation shows for computing the error signal of a layer, the gradient of the activation function is multiplied by other terms. Therefore, undefined gradient and zero gradient of the activation function pose a challenge in training neural networks as the weights are not updated. In quantization, where a constant piecewise and discrete function is commonly used, there exist regions with undefined and zero derivatives. The issue becomes more challenging in activation quantization, particularly when the activation function is non-differentiable. To address this problem, many works utilized the STE approach to estimate the gradient for non-differentiable functions.

## 4.1 Straight-Through Estimator

The idea of STE dates back to the 1950s when it was used for the perceptron learning algorithm [134]. Unlike the EBP algorithm, in the perceptron algorithm, it is not possible the propagation of the gradients from the last layer to the first layer. STE was raised in 2012 by Hinton again [135], who developed this method for training binary networks with multiple layers, where the activation function is non-differentiable. He assigned a gradient value of 1 to the positive arguments and zero otherwise, in the backward pass of the EBP algorithm. In 2013, Bengio *et al.* studied the challenge of estimating or propagating gradients through stochastic discrete neurons [136]. They suggested using the derivative of the S*igmoid* function as the STE in the backward pass. Since then, STE has been widely adopted in QAT.

In the training of a quantized network, STE is used for estimating the gradient of a non-differentiable quantized function in the backward pass. For model convergence during training, an accurate estimation of the quantization function is essential. Therefore, STE is an optimization problem as

$$\min_{\tilde{Q}(x)} E = |Q(x) - \tilde{Q}(x)| \tag{30}$$

In Equation (30), $Q(x)$ represents the forward activation function, while $\tilde{Q}(x)$ indicates its estimation in the backward pass.

**STE for the *Sign* function**: In the Bitwise Neural Networks method [106], the *Sign* function as Equation (3) is employed for the weights and activations binarization. Since the derivative of the *Sign* function is the *Impulse* function, which is non-differentiable, the STE approach is applied in the backward pass, where a value of 1 is assigned to the gradient of the *Sign* function. Figure 11a demonstrates the *Sign* function and its derivative [25].

In the Bi-RealNet method [25], three differentiable functions are investigated for the STE of the *Sign* function. Figure 11 presents the estimated functions and their derivative [25]. The first function, in Figure 11b, is *hard tanh* (*Clip*), which has been commonly used in previous methods such as the XNOR-Net [22] and Binarized Neural Networks (BNN) [42] methods as the STE for the *Sign* function. In *hard tanh*, the gradient is equal to 1 for values in the range [-1, 1], and it is equal to 0 for values outside this interval. The gradient of the *hard tanh* function is presented in Equation (31).

$$h'(x) = \begin{cases} 1 & |x| \leq 1 \\ 0 & |x| > 1 \end{cases} \tag{31}$$

Bi-RealNet introduced a new function, similar to *hard tanh* for STE of the *Sign* function, as shown in Figure 11c. Equation (32) shows this function, and Equation (33) represents its gradient.

$$F(a_r) = \begin{cases} -1 & a_r < -1 \\ 2a_r + a_r^2 & -1 \leq a_r < 0 \\ 2a_r - a_r^2 & 0 \leq a_r < 1 \\ 1 & a_r \geq 1 \end{cases} \tag{32}$$

$$\frac{\partial F(a_r)}{\partial a_r} = \begin{cases} 2 + 2a_r & -1 \leq a_r < 0 \\ 2 - 2a_r & 0 \leq a_r < 1 \\ 0 & o.w. \end{cases} \tag{33}$$

The shaded areas in Figure 11 indicate the difference between the *Sign* function and the estimated function. The shaded area for *hard tanh* is equal to 1, while for the proposed function in Figure 11c, it is equal to 2/3. It means that the function in Figure 11c is a more accurate estimator for the *Sign* function. As the order of the function in Equation (32) increases, the difference between the estimator and the *Sign* function decreases. Figure 11d illustrates the estimated function with a third-order, and Equation (34) determines its relation. The authors of the Bi-RealNet paper concluded that higher-order functions require more complex computations, and thus, the second-order function is acceptable.

$$F(a_r) = \begin{cases} -1 & a_r < -1 \\ (a_r + 1)^3 - 1 & -1 \leq a_r < 0 \\ (a_r - 1)^3 + 1 & 0 \leq a_r < 1 \\ 1 & a_r \geq 1 \end{cases} \quad (34)$$

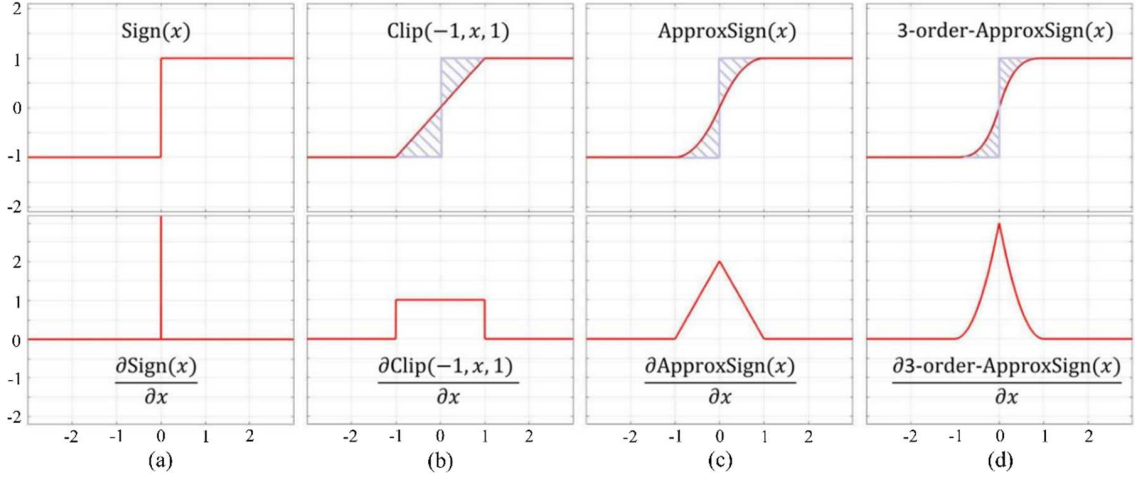

Figure 11: Estimators for the *Sign* function a) The *Sign* function and its derivative b) The hard tanh function and its derivative c) A second-order estimator function for the *Sign* function and its derivative d) A third-order estimator function for the *Sign* function and its derivative [25]

**Error Decay Estimator**: The Information Retention Network (IR-Net) method [125] proposed Error Decay Estimator (EDE) for estimating the *Sign* function in the backward pass as

$$F(x) = k \tanh(tx) \quad (35)$$

where $k$ and $t$ are computed as

$$t = T_{min} 10^{\frac{i}{N} \times log\frac{T_{max}}{T_{min}}}, \quad k = \max(1/t, 1) \quad (36)$$

In Equation (36), $i$ determines the epoch number in $N$ epochs, $T_{min}$ is set to $10^{-1}$, and $T_{max}$ is set to 10. EDE lies between the *identity* function ($y=x$) and *hard tanh* function. The *hard tanh* is close to the *Sign* function, but it discards the parameters outside the range [-1,1]. Consequently, those parameters are not updated anymore, leading to a loss of information. On the other hand, the *identity* function covers the parameters outside [-1,1] but has a significant difference from the *Sign* function, as indicated by the shaded area in Figure 12. EDE makes a trade-off between the *identity* and *hard tanh* functions by varying parameters $k$ and $t$ during training. Initially, $k$ is bigger than 1, making EDE closer to the *identity* function. As the number of epochs increases, $k$ gradually tends towards 1, causing EDE transition to *hard tanh* for achieving more accurate estimation.

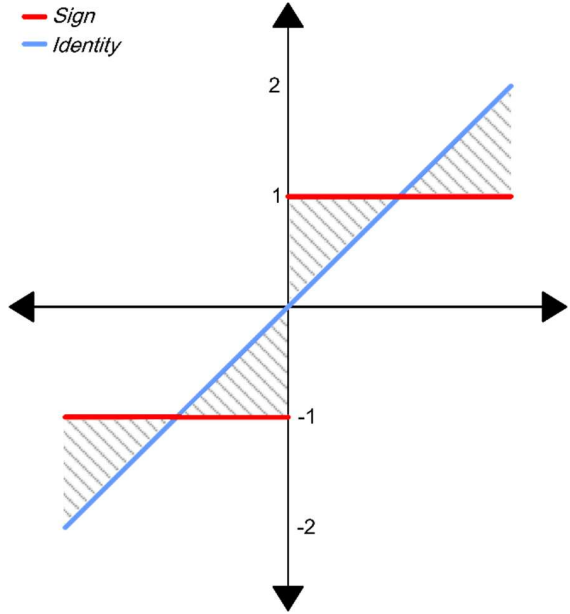

Figure 12: Distance between the *Sign* and identity functions

**Quantized *ReLU* and STE:** Cai *et al.* suggested Half-Wave Gaussian Quantization (HWGQ) for activations quantization [116]. HWGQ is a half-wave rectifier like *ReLU* known as quantized *ReLU* [137].

$$Q(x) = \begin{cases} q_i & x \in (t_i, t_{i+1}] \\ 0 & x \leq 0 \end{cases} \quad i = 1, \dots, m \quad q_i, t_i \in \mathbb{R}^+ \quad t_1 = 0, t_{m+1} = \infty \tag{37}$$

In Equation (37), $q_i$ represents the *i*th quantization level. Figure 13a shows the HWGQ function with quantization levels {0, $q_1$, $q_2$} [116]. Finding the optimal values of $q_i$s and $t_i$s is the main challenge in the HWGQ function, which depends on the distribution of full-precision values. The value $q_m$ is chosen in such a way that values greater than it, are considered negligible. The derivative of HWGQ is zero, and hence, Cai *et al.* examined three functions as estimators in the backward pass: 1) *Vanilla ReLU*, 2) *Clipped ReLU*, and 3) *Log-Tailed ReLU*.

The *ReLU* function is used as an estimator of HWGQ in the backward pass that is denoted as *Vanilla ReLU*. The derivative of *Vanilla ReLU* is presented in Equation (38). HWGQ is bounded to $q_m$ for *x*>0, whereas *Vanilla ReLU* tends to infinity, as demonstrated in Figure 13 [116]. This creates a mismatch between the two functions in the interval ($t_m$, ∞). Consequently, using *ReLU* in the backward pass leads to inaccurate gradients and unstable learning during training. The mismatch between HWGQ and *Vanilla ReLU* has a greater impact on deeper networks.

$$\tilde{Q}' = \begin{cases} 1 & x > 0 \\ 0 & o.w. \end{cases} \tag{38}$$

The second function is *Clipped ReLU* for estimating HWGQ in the backward pass as shown in Figure 13c. *Clipped ReLU* and its derivative are demonstrated in Equation (39). In *Clipped ReLU*, the weak point of the *Vanilla ReLU* is modified by setting the gradient to zero for *x*≥$q_m$. This modification makes *Clipped ReLU* a suitable estimator for HWGQ. The idea behind this modification comes from the fact that the frequency of the large values is commonly low, and these values are interpreted as outliers.

$$\tilde{Q} = \begin{cases} q_m & x > q_m \\ x & 0 < x \le q_m, \\ 0 & o.w. \end{cases} \quad \tilde{Q}' = \begin{cases} 1 & 0 < x \le q_m \\ 0 & o.w. \end{cases} \quad (39)$$

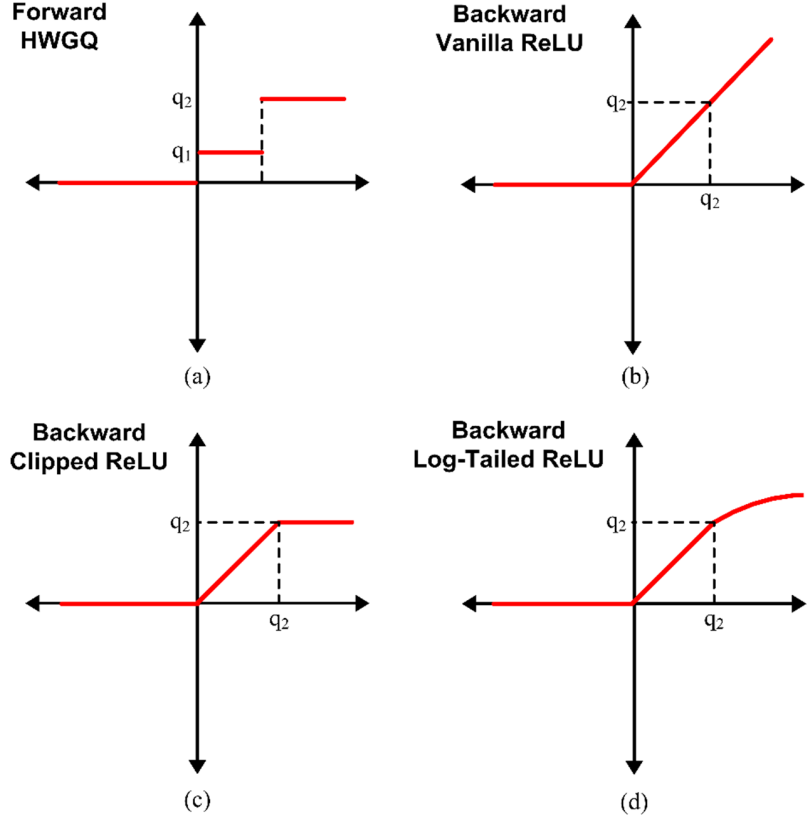

Figure 13: Estimators for HWGQ function a) Forward HWGQ with two positive quantization levels and zero level b) Vanilla ReLU c) Clipped ReLU d) Log-Tailed ReLU [116]

The third estimator function for HWGQ is *Log-Tailed ReLU*, which lies between *Vanilla ReLU* and *Clipped ReLU*, as illustrated in Figure 13d. The *Log-Tailed ReLU* function and its derivative are presented in Equations (40) and (41), respectively. Unlike *Clipped ReLU*, *Log-Tailed ReLU* does not completely ignore values $x > q_m$. Instead, the gradient of values $x > q_m$ gradually approaches zero as $x$ tends to infinity. Experimental results in the HWGQ method [116] show that *Log-Tailed ReLU* achieves higher accuracy in AlexNet compared to *Clipped ReLU*. However, *Clipped ReLU* achieves superior performance compared to *Log-Tailed ReLU* in VGGNet-Variant and ResNet-18, which are deeper than AlexNet.

$$\tilde{Q} = \begin{cases} q_m + \log(x - \tau) & x > q_m \\ x & 0 < x \le q_m, \quad \tau = q_{m-1} \\ 0 & x \le 0 \end{cases} \quad (40)$$

$$\tilde{Q}' = \begin{cases} 1/(x-\tau) & x > q_m \\ 1 & 0 < x \leq q_m, \quad \tau = q_{m-1} \\ 0 & x \leq 0 \end{cases} \tag{41}$$

**Bounded rectifier and STE**: In the ABC method [117], a bounded rectifier activation function was proposed for mapping the full-precision activations to the range [0,1] as

$$h_v(x) = clip(x+v, 0, 1) = \begin{cases} 1 & x+v > 1 \\ x+v & 0 < x+v < 1 \\ 0 & x+v < 0 \end{cases} \tag{42}$$

In Equation (42), $v$ represents a trainable shift parameter. After mapping the values to the range [0,1], Equation (43) is utilized for binarization as

$$A = H_v(R) = 2\mathbb{I}_{h_v(R) \geq 0.5} - 1 = \begin{cases} +1 & h_v \geq 0.5 \\ -1 & h_v < 0.5 \end{cases} \tag{43}$$

In Equation (43), $\mathbb{I}$ indicates the *Indicator* function. In the backward pass, STE is applied, and the gradient is computed as

$$\frac{\partial c}{\partial R} = \frac{\partial c}{\partial A} \mathbb{I}_{0 \leq R-v \leq 1} \tag{44}$$

**Parametric quantization function and STE**: Choi *et al.* [107] introduced a parametric clipped activation (PACT) function as

$$y = PACT(x) = 0.5(|x| - |x-\alpha| + \alpha) = \begin{cases} 0 & x \in (-\infty, 0) \\ x & x \in [0, \alpha) \\ a & x \in [\alpha, +\infty) \end{cases} \tag{45}$$

The PACT function maps the full-precision activations to the range [0, $\alpha$]. Then the output of the PACT function is quantized to $k$ bit-width precision using Equation (46).

$$y_q = round(y \frac{2^k - 1}{\alpha}) \frac{\alpha}{2^k - 1} \tag{46}$$

If $\alpha$ is equal to 1, the PACT function corresponds to the bounded rectifier function with $v$=0 in the ABC method [117]. The optimum $\alpha$ is found during training for minimizing the accuracy drop in quantization. It should be noted that the optimum value varies across different layers and models. Since Equation (46) is not differentiable, STE is employed for updating $\alpha$ as

$$\frac{\partial y_q}{\partial \alpha} = \frac{\partial y_q}{\partial y} \frac{\partial y}{\partial \alpha} = \begin{cases} 0 & x \in (-\infty, \alpha) \\ 1 & x \in [\alpha, +\infty) \end{cases} \tag{47}$$

The training is dependent on the value of $\alpha$. If the initial value of $\alpha$ is too small, according to Equation (47), most activations will fall in the range of non-zero gradient, causing frequent changes in the value of $\alpha$ during training and leading to low model accuracy. On the other hand, if the initial value of $\alpha$ is too large, the gradient will be zero for the majority of activations, leading to small gradients and the risk of gradient vanishing in the EBP algorithm. To address this, $\alpha$ is initialized with a large value that is not excessively large, and then reduced using L2-norm regularization.

Although the effectiveness of STE has been demonstrated in practice through the results of previous works, there is still a concern regarding the lack of theoretical proof for its performance. Therefore, in recent years, some researchers have made efforts to theoretically justify the performance of STE [138, 139].

Table 3 summarizes the forward quantization function and its estimator in the backward pass using STE for several previous works.

Table 3: Forward activation functions and their estimator in backward pass using STE in several methods

| Method | Forward activation function | Estimator in backward pass |
| --- | --- | --- |
| Bitwise Neural Networks [106] | $q = Sign(r)$ | $g_r = g_q$ |
| BNN [42] | $q = Sign(r)$ | $g_r = g_q \mathbb{I}_{\|r\| \leq 1}$ |
| XNOR-Net [22] | $q = Sign(r)$ | $g_r = g_q \mathbb{I}_{\|r\| \leq 1}$ |
| Dorefa-net [23] | $q = Sign(r)$ | $g_r = g_q$ |
| QNN [78] | $q = Sign(r)$ | $g_r = g_q \mathbb{I}_{\|r\| \leq 1}$ |
| HWGQ [116] | $q = \begin{cases} q_i & r \in (t_i, t_{i+1}] \\ 0 & r \leq 0 \end{cases}$ | $g_r = g_q \mathbb{I}_{0 < r \leq q_m}$ |
| ABC [117] | $q = 2\mathbb{I}_{x \geq 0.5} - 1, \quad x \in [0,1]$ | $g_r = g_q \mathbb{I}_{0 \leq r - v \leq 1}, \ r \in \mathbb{R}$ |
| Balanced quantization [79] | $q = \dfrac{round\big((2^k - 1)x\big)}{2^k - 1}, \ x \in [0,1]$ | $g_r = g_q$ |
| PACT [107] | $q = \dfrac{round\left(x \dfrac{2^k - 1}{\alpha}\right)\alpha}{2^k - 1}, \ x \in [0, \alpha]$ | $g_r = g_q \mathbb{I}_{r \geq \alpha}, \ r \in \mathbb{R}$ |
| Bi-RealNet [25] | $q = Sign(r)$ | $g_r = g_q \begin{cases} 2 + 2a_r & -1 \leq a_r < 0 \\ 2 - 2a_r & 0 \leq a_r < 1 \\ 0 & otherwise \end{cases}$ |
| IR-Net [125] | $q = Sign(r)$ | $g_r = g_q kt \mathbb{I}_{\|r\| \leq 1}$ $k$ and $t$ are defined in Equation (36) |

## 4.2 Weights Update in Quantized Network Training

After the weight quantization, the weights are limited to discrete quantization levels. There is a distance between the quantization levels equal to the step sizes. During the update step of the EBP algorithm, the change in weights may be smaller than the quantization step sizes. As a result, when the quantized weights are used in the update relation during the training phase, they remain unaltered, as shown in Figure 14a. In Figure 14, suppose that q₁, q₂, and q₃ are the quantization levels. $w_f^t$ and $w_f^{t+1}$ represent the full-precision weights before and after the update, respectively, while $w_q^t$ and $w_q^{t+1}$ indicate their corresponding values after quantization. The relations related to Figure 14a are as follows:

$$w_q^t = quantize(w_f^t) = q_1$$

$$w_f^{t+1} = w_q^t - \gamma \frac{\partial L}{\partial w}$$

$$w_q^{t+1} = quantize(w_f^{t+1}) = q_1 \Longrightarrow w_q^t = w_q^{t+1} \Longrightarrow non-update \tag{48}$$

In Equation (48), $L$ represents the loss function, $\gamma$ is the learning rate, and *quantize*() denotes the quantization function. In Figure 14a and Equation (48), the value of the quantized weight before updating is equal to q₁. During weight update, due to the small change in the weight, the updated value is again quantized to q₁, resulting in no actual weight update. To address this issue, most works keep the full-precision weights after quantization and utilize them in the update step. The full-precision weights are used during the update step, and the updated weights are quantized. Figure 14b and Equation (49) present the update step using full-precision weights. As seen, the weight value is updated from q₁ to q₂.

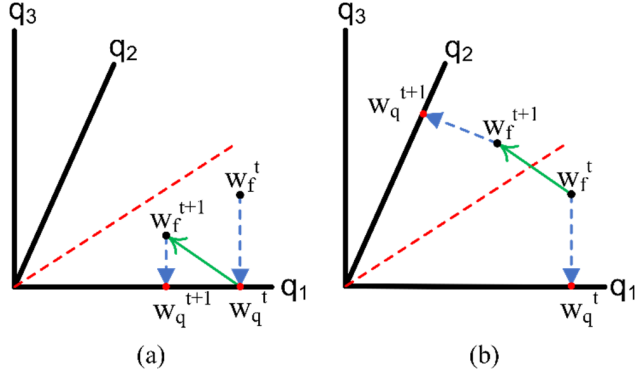

Figure 14: Update step in weight quantization a) Updating using the quantized weights b) Updating using the full-precision weights

$$w_q^t = quantize(w_f^t) = q_1$$

$$w_f^{t+1} = w_f^t - \gamma \frac{\partial L}{\partial w}$$

$$w_q^{t+1} = quantize(w_f^{t+1}) = q_2 \Longrightarrow w_q^t \neq w_q^{t+1} \Longrightarrow update \tag{49}$$

### 4.3 Some Effective Parameters in Training

Due to the discretization in quantization, training in a quantized network is challenging and requires new methods to improve model accuracy. In a discrete space, the parameter space becomes significantly smaller, which makes convergence in the training algorithm more challenging. This problem is more considerable in lower bit-width quantization, such as binary and ternary quantization. Therefore, new techniques are required for training quantized neural networks. In this section, we discuss the role of learning rate, network structure, and regularization in the training of quantized neural networks.

#### 4.3.1 Learning Rate

The initial value of the learning rate has a considerable effect on neural network training. In quantized networks, a small learning rate often yields better. Using a large learning rate causes weight values to frequently jumping among the limited quantization levels. This instability during training damages the convergence of the learning model. Work [115] discussed the effect of the learning rate in the training of a binarized neural network.

#### 4.3.2 Network Structure

Sometimes structural adjustments are necessary for the neural network after quantization. For instance, the max-pooling layer in some binary DNNs is displaced. In DCNNs, a max-pooling layer commonly comes immediately after the activation layer. However, in a binary neural network, where the *Sign* function is used for the binarization of the activations, placing the max-pooling layer immediately after the *Sign* function result in an output matrix containing only +1 elements as the values in a binarized matrix are -1 and +1. This leads to a loss of information. To address this issue, works [22, 117] have rearranged the max-pooling layer after the convolution layer.

*4.3.3 Regularization*

In DNNs, regularization is commonly applied to prevent overfitting, which is more likely to occur due to the huge number of parameters. L1 and L2 regularization are frequently employed in full-precision networks to push network parameters toward zero and reduce their impact. However, in quantization, approximating weights with low bit-width precision acts as a regularizer, and pushing the weights toward zeros can lead to a significant drop in accuracy. To solve this problem, some previous works have proposed new regularization approaches compatible with their quantization method [115, 140-143]. Bit-level Sparsity Quantization (BSQ) [140] suggested a regularization for mixed-precision quantization. Another work [141] offered a periodic regularization to push the full-precision weights toward the quantization levels for increasing the quantization accuracy. ProxQuant [142] proposed a regularization for pushing the full-precision weights to the binary quantization levels.

Tang *et al.* [115] introduced a new regularization for binary quantization, which encourages weights to approach quantization levels {-1,1} in order to minimize quantization error. Equation (50) demonstrates this regularization.

$$J(W, b) = L(W, b) + \lambda \sum_{l=1}^{L} \sum_{i=1}^{N_l} \sum_{j=1}^{M_l} (1 - (W_{l,ij})^2) \qquad (50)$$

In Equation (50), $L(W, b)$ represents the loss function, and the second term denotes the regularization relation. $L$ indicates the number of layers. $N_l$ and $M_l$ are the dimensions of the weight matrix in layer $l$. The parameter $\lambda$ controls the effect of the loss function and regularization term.

The Smart Quantization (SQ) method [127] conducted quantization in 1-bit and 2-bit precision and introduced an adaptive binary-ternary regularization. For binary quantization, the quantization levels are {-$\alpha$, +$\alpha$}, and for ternary quantization, the quantization levels are {-$\alpha$, 0, +$\alpha$}, where $\alpha$ represents the scale factor. The adaptive binary-ternary regularization is defined as

$$R(W, \alpha, \beta) = \min\left(||w| + \mu|^p, ||w| - \mu|^p, \tan(\beta)|w|^p\right) \qquad (51)$$

In Equation (51), $w$ indicates the full-precision weight, and $p$ denotes the order of the regularization function and is set as 1. $\beta \in (\frac{\pi}{4}, \frac{\pi}{2})$ controls the transition between binary and ternary quantization. As $\beta$ approaches $\frac{\pi}{2}$, the value of $tan(\beta)$ increases and moves away from zero, resulting in a transition to binary regularization in Equation (51). On the other hand, as $\beta$ approaches $\frac{\pi}{4}$, the regularization converges to ternary quantization. Another work [143] defined the quantization regularization as

$$R = \sum_n^N \sum_i^{card(W_n)} \frac{|W_{n_i} - W_{q_{n_i}}|}{\max(Q_n) \times card(W_n)} \qquad (52)$$

In Equation (52), $N$, $W$, and $Wq$ represent the number of layers, full-precision weights, and quantization weights, respectively. $card(Wn)$ determines the number of weights in layer $n$. This regularization is defined as the mean of the absolute difference between the full-precision weights and their corresponding quantization levels. Equation (52) was extended to Equation (53) for POT quantization as

$$R = \sum_n^N \sum_i^{card(W_n)} \frac{|W_{n_i} - W_{q_{n_i}}||W_{n_i}|}{\max(Q_n)^2 \times card(W_n)} \qquad (53)$$

In the POT quantization, the density of quantization levels decreases as the values become larger, as shown in Figure 6c. In that respect, Equation (53) applies stronger regularization to larger weights.

In general, in all quantization regularization methods, the full-precision weights are gradually pushed through the quantization levels during training for alleviating accuracy degradation resulting from quantization.

## 5 OPERATIONS IN QUANTIZATION

The main operation in DCNNs is the MAC operation, which is performed in convolution and FC layers. MAC is a dot product between weights and inputs as

$$y = W \odot X \qquad W \in \mathbb{R}^n, X \in \mathbb{R}^n \tag{54}$$

In Equation (54), $W$ and $X$ represent the weight and input matrices, respectively, and $\odot$ indicates the dot product. Figure 15 demonstrates an example of the MAC operation in a convolution layer, where a dot product is performed between a filter and several patches of the input feature map for producing the output feature map. As presented in Table 1, DCNNs involve a considerable number of MACs that require floating-point multiplications and additions. These floating-point operations (FLOPs) create a bottleneck in the implementation of accelerators for DCNNs. After quantization, 32-bit floating-point numbers are mapped to values with lower bit-widths, such as 8, 4, 2, and 1 bit-width. This mapping allows for the replacement of the floating-point operations with more efficient integer or bitwise ones on hardware platforms. In the following, the proposed methods for efficiently implementing the MAC operation in quantized DCNNs are discussed.

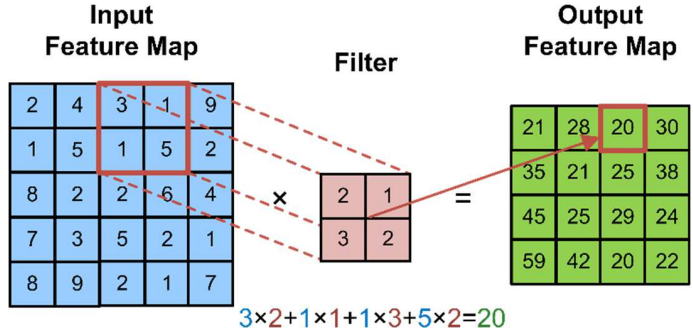

Figure 15: The dot product of an input feature map with a filter and producing output feature map

Miyashita *et al.* [110] developed two methods based on logarithmic quantization. In the first method, only activations were quantized in the base-2 logarithm, and the MAC operation was performed as

$$y = W \odot X \simeq \sum_{i=1}^{n} w_i \times 2^{\tilde{x}_i} = \sum_{i=1}^{n} Bitshift(w_i, \tilde{x}_i) \tag{55}$$

In Equation (55), $w_i$ represents a fixed-point number, and $\tilde{x}_i$ denotes an integer number computed as

$$\tilde{x}_i = Quantize(log_2(x_i)) \tag{56}$$

In Equation (55), the high-cost floating-point multiplication is replaced by a simple bit-shift operation, where $w_i$ shifts by $\tilde{x}_i$ in fixed-point arithmetic. In the second method proposed by Miyashita *et al.*, the weights and activations are quantized using base-2 logarithm quantization. Subsequently, the MAC operation is modified as Equation (57), where the floating-point multiplications are replaced with bit-shift operations and integer additions.

$$y = W \odot X \simeq \sum_{i=1}^{n} 2^{\widetilde{w}_i} \times 2^{\widetilde{x}_i} = \sum_{i=1}^{n} 2^{\widetilde{w}_i + \widetilde{x}_i} = \sum_{i=1}^{n} Bitshift(1, \widetilde{w}_i + \widetilde{x}_i) \quad (57)$$

Works [123, 144] employed POT quantization, which allows using bit-shift operation in the MAC operation. When the both weights and activations are quantized to fixed-point integer, the bitwise operations can be applied for MAC. Suppose that weights (*W*) are represented as *M*-bit integer and activations (*X*) are quantized to *N*-bit integer:

$$W = \sum_{m=0}^{M-1} w_m 2^m \ , X = \sum_{n=0}^{N-1} x_n 2^n, \forall m, n, w_m, x_n \in \{0,1\} \quad (58)$$

The MAC operation between *W* and *X* can be performed using bitwise operations as

$$y = W \odot X = \sum_{m=0}^{M-1} \sum_{n=0}^{N-1} 2^{m+k} bitcount[and(w_m, x_n)] \quad (59)$$

The Dorefa-net [23] and Balanced quantization [79] methods applied Equation (59) for performing the MAC operation. In a binary neural network, where both weights and activations are quantized to 1-bit, the MAC operation can be efficiently performed using bitwise operations as

$$y = W \odot X = bitcount(xnor(W, X)) \quad (60)$$

In Equation (60), *W* and *X* indicate binary weights and activations matrices with values of {-1,1}, respectively. The XNOR operation is used instead of floating-point multiplication, while the *bitcount* function counts the number of 1s in a bit string, replacing the floating-point additions. If binary quantization is used with values {0,1} instead of {-1,1}, the XNOR operation is replaced by the AND operation. XNOR-Net [22], BNN [42], Bitwise Neural Networks [106], and [116] employed Equation (60) for the MAC operation.

Zhang *et al.* [122] performed weights and activations quantization using Equation (7) with $K_w$-bit and $K_a$-bit precisions, respectively. They utilized bitwise operations for computing MAC between filters and activations as

$$y = W \odot X = Q(W, v^w) Q(X, v^a) = \sum_{i=1}^{K_w} \sum_{j=1}^{K_a} v_i^w v_j^a \ bitcount(xnor(W, X)) \quad (61)$$

In Equation (61), $v^m \in \mathbb{R}^{K_w}$ and $v^a \in \mathbb{R}^{K_a}$ represent the learnable floating-point basis vectors for weights and activations, respectively, as defined in Equation (7).

In quantization methods in which the zero level is defined, the majority of the full-precision numbers are mapped to zero. When either the weight or activation operand is zero, the multiplication operation is omitted, leading to a further reduction in computations. Moreover, the activations smaller than zero are set to zero using the *ReLU* function. For instance, Venkatesh *et al.* [128] demonstrated that only 16% of operations in the forward pass and 33% of operations in the backward pass are non-zero when using ternary weight quantization in the training of Resnet-34 on ImageNet.

## 6 LAYERS IN QUANTIZATION

Quantization of each Layer in DCNNs leads to different effects on prediction accuracy, which is influenced by the structure and position of the layer within the network. In FC layers, each weight is connected to only one input, whereas in convolution layers, weights are shared among multiple inputs. Consequently, it seems that the quantization of weights in convolution layers has a greater impact on the prediction accuracy drop compared to FC layers. Moreover, the position of layers within a DCNN also influences the efficiency of quantization. The size of the input channels and output channels in the last convolution layers in the DCNNs are generally larger than those in the first convolution layers. As a result, the last layers contain most of the parameters with a wide range of values, and quantization of them leads to more significant efficiency in memory saving and possessing speed-up.

## 6.1 First and Last Layers

Among all the layers of a DCNN, the first and last layers cause more accuracy drop in quantization. Input and output layers are directly related to the input and output of the network, respectively. The first convolution layer is more sensitive compared to other layers since it receives the main data as input. The filters in the first layer must extract the most important and informative features. In DCNNs, the output of each layer relies on the combination of the extracted features from the previous layers. If the first layer fails to extract informative features, this will propagate layer by layer, resulting in a decrease in accuracy. On the other hand, the last layer is directly related to the output, as its output is compared with the desired result, while quantization leads to significant variations in the outputs [115].

Besides, the first layer has fewer weights and MAC operations compared to other layers due to its smaller size of input and output channels. As a result, the quantization of the first layer has a relatively small effect on the compression rate and the processing speed-up. Similarly, the quantization of the last layer, which is typically an FC layer in most DCNNs, has a minor impact on the processing speed-up due to the relatively small number of operations involved.

BRECQ [145] has compared the effect of quantization of the first and last layers in terms of memory saving and latency. This study has revealed that quantization of the last FC layer is more efficient in memory saving compared to the first layer, as the last layer typically involves more weights. In terms of latency, it depends on the architecture of DCNN. In some cases, quantization of the first layer leads to more efficient processing speed-up compared to the last layer.

Some previous works kept the first and last layer in full precision [22, 23, 24, 113, 146-149], while there are also studies that employed quantization with higher precision for the first and last layers compared to the hidden layers [24, 42, 150-152]. As an example, in work [42], the activations of all layers were quantized to 1-bit precision, except the input of the first layer, which was represented in 8-bit fixed-point.

Some researchers proposed efficient solutions for achieving greater compression through the quantization of the first and last layers while maintaining model accuracy. Studies such as [24, 115] offered a solution for the efficient quantization of weights and activations in the last layer. These works inserted a scale layer at the end of a DCNN, and the output of the last layer is multiplied by a scale parameter to harmonize it with the desired output. The scale parameter, which is a learnable parameter, converges to an optimum value during training. By using the scale layer and quantization of the last layer, work [115] achieved a compression rate of approximately 4.5 times in NIN-net [153] compared to keeping the last layer in full precision while top-5 accuracy decreased only %1.

## 6.2 Mixed-Precision

Recently, there have been suggestions for mixed-precision quantization according to the sensitivity of the layers in the network, in which each layer is assigned a different bit-width. The sensitivity of a layer refers to the impact it has on prediction accuracy when represented with low precision. The mixed-precision quantization is an optimization problem with two objectives: accuracy and compression rates, aiming to find the optimum tradeoff between them.. The search space for mixed-precision quantization grows exponentially with the number of layers. Suppose that there are $M$ possible bit precisions for $N$ layers, the time complexity of finding the optimum precision is $O(M^N)$, which makes it a challenging task. Works [70, 71, 140, 148, 150, 154-162] proposed mixed-precision quantization.

HAWQ [154] performs mixed-precision quantization in multiple stages, where in each stage, a subset of layers in the network is quantized. The bit-width of each layer is determined based on its Hessian spectrum and the number of parameters. Layers with a smaller Hessian spectrum are found to be less sensitive to quantization, while layers with a large number of parameters can be quantized in lower bit-width, leading to higher compression rates. Therefore, HAWQ conducts quantization in lower precision for layers with a smaller Hessian spectrum and a large number of parameters.

The Bit-level Sparsity Quantization (BSQ) method [140] proposed a regularizer based on the group Lasso [163] for performing mixed-precision quantization during training. Another approach, the Differentiable Neural Architecture Search (DNAS) method [148], introduces a loss function for making a trade-off between accuracy and compression rate in mixed-precision quantization as

$$L(a, w_a) = CrossEntropy(a) \times C(Cost(a)) \tag{62}$$

In Equation (62), $a$ represents model architecture, and *Cost*() is a function that determines the architecture cost. $C$ indicates a weighting function that controls the trade-off between the cross entropy and the cost term. Model cost is computed based on the number of parameters, FLOPs, and the bit-width of weights and activations.

Work [159] employed an iterative approach for mixed-precision quantization. It ranks the layers based on the accuracy gain, and for estimating the accuracy gain of each layer, a proxy classifier is inserted after that layer. The layers with lower accuracy gain are quantized first, and after assigning the bit-width for all layers, a fin-tunning is performed.

Some works suggested channel-wise mixed-precision quantization [126, 164-166]. In general, the optimum precision for different parts of a neural network is dependent on the network architecture and dataset. The implementation of mixed-precision quantization is more complex compared to fixed-precision quantization in both software and hardware aspects [152].

## 7 EVALUATION AND DISCUSSION

To evaluate the quantization methods, various metrics are commonly reported, including model accuracy, compression rate, memory consumption, FLOPs, energy consumption, and processing speed [17, 25, 42, 67, 68, 71, 77, 109, 115, 118, 132, 140, 146, 149-151, 154, 155, 158, 159, 167-170].

Quantization reduces memory usage by storing values in low bit-width, and it also decreases FLOPs by replacing high-cost floating-point operations with low-cost simple operations. Representing values in low bit-width allows for storing the model on energy-efficient on-chip memory instead of power-hungry off-chip DRAM. Using on-chip memories and employing low-cost operations leads to power saving and processing speed-up in the implementation of the quantized models on hardware platforms. However, the accuracy commonly decreases after quantization due to information loss resulting from the limited quantization levels and elimination of larger weights that have a greater impact on accuracy. While lower bit-width quantization offers benefits such as memory saving and processing speed-up, it often results in more accuracy degradation. Therefore, a trade-off between compression rate and model accuracy is necessary.

### 7.1 Datasets

For the evaluation of DCNNs in the image classification task, the common benchmark datasets are:

1. **MINST** [102] is a dataset of handwritten digits with 10 classes, each representing the digits from 0 to 9.
2. **CIFAR-10** [171] includes 10 different classes with 6000 images per class.
3. **SVHN** [172] is a dataset of real-world digits.

4. **ImageNet** (ILSVRC12) [173, 174] is a large-scale dataset of natural images with high resolution. Images are represented in 1000 categories and normally resized to 224×224 before feeding to the network.

Table 4 presents additional specifications of these datasets.

Table 4: Specifications of some datasets in image classification

| Dataset | Training Samples | Test Samples | Type of Images | Size of Images |
|---|---|---|---|---|
| **MNIST** | 60K | 10K | Gray-scale | 28×28 |
| **CIFAR-10** | 50K | 10K | Color (RGB) | 32×32 |
| **SVHN** | 604K | 26K | Color (RGB) | 32×32 |
| **ImageNet** | About 1.2M | 100K | Color (RGB) | - |

## 7.2 Accuracy on Small Datasets

MNIST, CIFAR-10, and SVHN are relatively small datasets in comparison with ImageNet, and models achieve higher accuracy on them. CIFAR-10 is more frequently used than MNIST and SVHN in experiments. Table 5 demonstrates the accuracy of SOA methods on CIFAR-10. The reported results are directly presented from the original papers. It is important to note that the situation of training for a full-precision network is different in various works. For example, they used different learning rates, initial values, and frameworks. Consequently, the reported accuracy for the full-precision model in previous works is different. In Table 5, we present the minimum and maximum reported accuracy for the full-precision models.

Table 5: Accuracy of some SOA methods on CIFAR-10 with various bit-width precision

| Network | Method | Bit-Width (W/A) | Accuracy (%) |
|---|---|---|---|
| **VGGNet-Small [21]** | Full-Precision | 32/32 | 91.7 to 93.8 |
| | BinaryConnect [21] | 1/32 | 91.73 |
| | BNN [42] | 1/1 | 89.85 |
| | HWGQ [116] | 1/2 | 92.51 |
| | LQ-Nets [122] | 3/2 | 93.8 |
| | | 2/32 | 93.8 |
| | | 2/2 | 93.5 |
| | | 1/32 | 93.5 |
| | | 1/2 | 93.4 |
| | RBNN [175] | 1/1 | 91.3 |
| | IR-Net [125] | 1/1 | 90.4 |
| | DGRL [169] | 1/1 | 92.62 |
| | DMBQ [126] | 0.7/32 | 93.7 |
| | | 1/2 | 93.9 |
| **VGGNet7-128 [112]** | Full-Precision | 32/32 | 92.88 to 93 |
| | TWN [112] | 2/32 | 92.56 |
| | DSQ [147] | Mixed | 91.54 |
| **VGGNet-11 [11]** | Full-Precision | 32/32 | 91.93 to 92.13 |
| | BCGD [137] | 32/4 | 91.66 |
| | | 4/4 | 91.31 |
| | | 2/4 | 90 |
| | | 1/4 | 89.12 |
| | Yin *et al.* [176] | 2/32 | 91.01 |
| | | 1/32 | 89.28 |

| Network | Method | Bit-Width (W/A) | Accuracy (%) |
| --- | --- | --- | --- |
| **VGGNet-16 [11]** | Full-Precision | 32/32 | 93.59 |
| | Yin *et al.* [176] | 2/32 | 93.2 |
| | | 1/32 | 91.98 |
| **ResNet-18 [8]** | Full-Precision | 32/32 | 93 to 95.49 |
| | Yin *et al.* [176] | 2/32 | 94.98 |
| | | 1/32 | 94.19 |
| | RBNN [175] | 1/1 | 92.2 |
| | IR-Net [125] | 1/1 | 91.5 |
| | Liu *et al.* [159] | 3.07/4.35 | 94.11 |
| | | 2.7/3.11 | 94.02 |
| | | 1.96/2.55 | 93.42 |
| **ResNet-20 [8]** | Full-Precision | 32/32 | 91.25 to 92.96 |
| | TTQ [113] | 2/32 | 91.13 |
| | PACT [107] | 5/5 | 91.7 |
| | | 4/4 | 91.3 |
| | | 3/3 | 91.1 |
| | | 2/2 | 89.7 |
| | MLQ [120] | 2/32 | 90.02 |
| | ELQ [121] | 2/32 | 91.45 |
| | | 1/32 | 91.15 |
| | LQ-Nets [122] | 3/32 | 92 |
| | | 3/3 | 91.6 |
| | | 3/2 | 91.1 |
| | | 2/32 | 91.8 |
| | | 2/2 | 90.2 |
| | | 1/32 | 90.1 |
| | | 1/2 | 88.4 |
| | Yin *et al.* [176] | 2/32 | 90.07 |
| | | 1/32 | 87.82 |
| | ProxQuant [142] | 2/32 | 91.6 |
| | | 1/32 | 90.65 |
| | BCGD [137] | 32/4 | 91.66 |
| | | 4/4 | 91.65 |
| | | 2/4 | 90.75 |
| | | 1/4 | 89.98 |
| | HAWQ [154] | 2/4 | 92.22 |
| | RBNN [175] | 1/1 | 86.5 |
| | IR-Net [125] | 1/32 | 90.8 |
| | | 1/1 | 85.4 |
| | SLB [177] | 4/32 | 92.1 |
| | | 4/8 | 91.8 |
| | | 4/4 | 91.6 |
| | | 2/32 | 92 |
| | | 2/8 | 91.7 |
| | | 2/4 | 91.3 |
| | | 1/32 | 90.6 |
| | | 1/8 | 90.5 |
| | | 1/4 | 90.3 |
| | | 1/2 | 89.5 |

| Network | Method | Bit-Width (W/A) | Accuracy (%) |
|---|---|---|---|
| | DMBQ [126] | 1/1 | 85.5 |
| | | 2/32 | 92.5 |
| | | 1/32 | 91.4 |
| | | 2/2 | 91.7 |
| | | 1/2 | 90.4 |
| | Liu et al. [159] | 3.3/3.4 | 92.82 |
| | | 2.3/2.7 | 91.65 |
| | | 1.5/2.3 | 90.64 |
| **ResNet-32 [8]** | Full-Precision | 32/32 | 92.33 to 93.4 |
| | TTQ [113] | 2/32 | 91.37 |
| | Yin et al. [176] | 2/32 | 92.04 |
| | | 1/32 | 90.65 |
| | ProxQuant [142] | 2/32 | 92.35 |
| | | 1/32 | 91.47 |
| **ResNet-34 [8]** | Full-Precision | 32/32 | 95.7 |
| | Yin et al. [176] | 2/32 | 95.07 |
| | | 1/32 | 94.66 |
| **ResNet-44 [8]** | Full-Precision | 32/32 | 92.82 |
| | TTQ [113] | 2/32 | 92.98 |
| | ProxQuant [142] | 2/32 | 92.95 |
| | | 1/32 | 92.05 |
| **ResNet-56 [8]** | Full-Precision | 32/32 | 93.03 to 94.46 |
| | TTQ [113] | 2/32 | 93.56 |
| | ELQ [121] | 2/32 | 93.7 |
| | | 1/32 | 92.82 |
| | ProxQuant [142] | 1/32 | 92.3 |
| | Liu et al. [159] | 3.37/3.42 | 93.74 |
| | | 2.37/2.75 | 92.89 |
| | | 1.82/2.42 | 92.22 |
| **MobileNetV2** | Full-Precision | 32/32 | 94.24 |
| | Liu et al. [159] | 3.32/3.39 | 84.82 |
| | | 2.48/2.83 | 74.42 |
| | | 1.32/2.14 | 63.92 |

The results in Table 5 demonstrate that the majority of methods in low bit-width have achieved an accuracy close to the full-precision model and in some cases, better than it. For instance, DMBQ [126] has achieved an accuracy of 93.9% for VGGNet-Small with 1-bit weights and 2-bit activations quantization, surpassing the accuracy of the full-precision model. Additionally, TTQ [113] and ProxQuant [142] using ternary weight quantization, have achieved higher accuracy compared to the full-precision model for ResNet-44.

### 7.3 Accuracy on ImageNet

ImageNet is considered as a large-scale and challenging dataset, making it a crucial benchmark for evaluating the accuracy of DCNN models in the image classification task. Table 6 presents the accuracy of several SOA methods on ImageNet. The results are reported from the original papers. It is important to consider that the training of the full-precision networks was performed in different situations in various works. Therefore, in Table 6, we indicate the minimum and maximum reported accuracy for the full-precision models.

Table 6: TOP-1 and TOP-5 accuracy of some SOA methods on ImageNet with various bit-width precision

| Network | Method | Bit-width(W/A) | Top-1 (%) | Top-5 (%) |
|---|---|---|---|---|
| **AlexNet [1]** | Full-precision | 32/32 | 56.6 to 61.8 | 80.23 to 83.5 |
| | DeepCompression [17] | CONV:8, FC:5/32 | 57.22 | 80.3 |
| | | CONV:8, FC:4/32 | 57.21 | 80.27 |
| | | CONV:4, FC:2/32 | 55.23 | 77.67 |
| | Dorefa-net [23] | 1/4(G:32) | 53 | - |
| | | 1/4(G:6) | 48.2 | - |
| | | 1/3(G:32) | 48.4 | - |
| | | 1/3(G:6) | 47.1 | - |
| | | 1/2(G:32) | 49.8 | - |
| | | 1/2(G:6) | 46.1 | - |
| | | 1/1(G:32) | 43.6 | - |
| | | 1/1(G:6) | 39.5 | - |
| | TTQ [113] | 2/32 | 57.5 | 79.7 |
| | Miyashita *et al.* [110] | 32/4 | - | 77.6 |
| | | 32/3 | - | 77.1 |
| | QNN [78] | 1/2 | 51.03 | 73.67 |
| | LogNet [114] | 5/32 | - | 74.6 |
| | | 4/32 | - | 73.4 |
| | Tang *et al.* [115] | 1/2 | 46.6 | 71.7 |
| | HWGQ [116] | 1/32 | 52.4 | 75.9 |
| | | 1/2 | 52.7 | 76.3 |
| | INQ [118] | 5/32 | 57.39 | 80.46 |
| | FGQ [119] | 2/8 | 49.04 | - |
| | | 2/4 | 49 | - |
| | WQ [80] | 4/4 | 55.8 | - |
| | | 2/3 | 51.37 | 75.49 |
| | Balanced [79] | 2/2 | 55.7 | 78 |
| | Zuang *et al.* [24] | 4/4 | 58 | 81.1 |
| | | 2/2 | 51.6 | 76.2 |
| | PACT [107] | 32/4 | 55.5 | 77.6 |
| | | 32/3 | 55.6 | 77.8 |
| | | 32/2 | 54.9 | 77.2 |
| | | 3/3 | 55.6 | 78 |
| | | 3/2 | 54.6 | 77.1 |
| | | 2/3 | 55.4 | 77.9 |
| | | 2/2 | 55 | 77.7 |
| | SLQ [120] | 5/32 | 57.56 | 80.5 |
| | MLQ [120] | 2/32 | 54.24 | 77.78 |
| | ELQ [121] | 2/32 | 57.88 | 80.22 |
| | | 1/32 | 56.95 | 79.77 |
| | LQ-Nets [122] | 2/32 | 60.5 | 82.7 |
| | | 2/2 | 57.4 | 80.1 |
| | | 1/2 | 55.7 | 78.8 |
| | Yang *et al.* [178] | 2/32 | 60.9 | 83.2 |
| | | 1/32 | 58.8 | 81.7 |
| | | 1/2 | 55.4 | 78.8 |
| | | 1/1 | 47.9 | 72.5 |
| | KDE-KM [131] | 4/32 | 47.43 | 72.1 |

| Network | Method | Bit-width(W/A) | Top-1 (%) | Top-5 (%) |
|---|---|---|---|---|
| **VGGNet-16 [11]** | BitPruning [124] | 3.875/4.375 | 55.07 | - |
| | Full-precision | 32/32 | 68.54 to 71.55 | 88.65 to 90.33 |
| | DeepCompression [17] | CONV:8, FC:5/32 | 68.83 | 89.09 |
| | Miyashita *et al.* [110] | 32/4 | - | 89.8 |
| | | 32/3 | - | 89.2 |
| | LogNet [114] | 5/32 | - | 86 |
| | | 4/32 | - | 85.2 |
| | | 4/4 | - | 84.8 |
| | INQ [118] | 5/32 | 70.82 | 90.3 |
| | SLQ [120] | 5/32 | 72.23 | 91 |
| | | 4/32 | 71.18 | 90.25 |
| | | 3/32 | 68.38 | 88.55 |
| | KDE-KM [131] | 4/32 | 67.76 | 88.14 |
| **VGGNet-Variant [116]** | Full-precision | 32/32 | 69.8 | 89.3 |
| | HWGQ [116] | 1/2 | 64.1 | 85.6 |
| | LQ-Nets [122] | 2/2 | 68.8 | 88.6 |
| | | 1/2 | 67.1 | 87.6 |
| **GoogleNet [2]** | Full-precision | 32/32 | 68.7 to 71.6 | 88.9 to 91.2 |
| | BWN [22] | 1/32 | 65.5 | 86.1 |
| | QNN [78] | 6/6 (G:6) | 66.4 | 83.1 |
| | | 4/4 | 66.5 | 83.4 |
| | INQ [118] | 5/32 | 30.98 | 89.28 |
| | HWGQ [116] | 1/2 | 63 | 84.9 |
| | Balanced [79] | 4/4 | 67.7 | 87.3 |
| | SLQ [120] | 5/32 | 69.1 | 89.19 |
| | LQ-Nets [122] | 2/2 | 68.8 | 88.1 |
| | | 1/2 | 65.6 | 86.4 |
| **ResNet-18 [8]** | Full-precision | 32/32 | 69.3 to 71.8 | 89.2 to 89.6 |
| | TWN [112] | 2/32 | 61.8 | 84.2 |
| | BWN [22] | 1/32 | 60.8 | 83 |
| | XNOR-Net [22] | 1/1 | 51.2 | 73.2 |
| | TTQ [113] | 2/32 | 66.6 | 87.2 |
| | HWGQ [116] | 1/32 | 61.3 | 83.6 |
| | | 1/2 | 59.6 | 82.2 |
| | ABC [117] | 1/32 | 68.3 | 87.9 |
| | | 1/1 | 65 | 85.9 |
| | INQ [118] | 5/32 | 68.98 | 89.1 |
| | | 4/32 | 68.89 | 89.01 |
| | | 3/32 | 68.08 | 88.36 |
| | | 2/32 | 66.02 | 87.13 |
| | Balanced [79] | 2/2 | 59.4 | 82 |
| | PACT [107] | 32/4 | 70 | 89.3 |
| | | 32/3 | 69.2 | 88.9 |
| | | 32/2 | 67.5 | 87.6 |
| | | 3/3 | 68.1 | 88.2 |
| | | 2/2 | 64.4 | 85.6 |
| | | 1/32 | 65.8 | 86.7 |
| | | 1/3 | 65.3 | 85.9 |
| | | 1/2 | 62.9 | 84.7 |

| Network | Method | Bit-width(W/A) | Top-1 (%) | Top-5 (%) |
|---|---|---|---|---|
| | SLQ [120] | 5/32 | 69.09 | 89.15 |
| | ELQ [121] | 2/32 | 67.52 | 88.05 |
| | | 1/32 | 64.72 | 86.04 |
| | LQ-Nets [122] | 4/32 | 70 | 89.1 |
| | | 4/4 | 69.3 | 88.8 |
| | | 3/32 | 69.3 | 88.8 |
| | | 3/3 | 68.2 | 87.9 |
| | | 2/32 | 68 | 88 |
| | | 2/2 | 64.9 | 85.9 |
| | | 1/2 | 62.6 | 84.3 |
| | Yin *et al.* [176] | 2/32 | 66.5 | 87.3 |
| | | 1/32 | 63.2 | 85.1 |
| | BCGD [137] | 4/8 | 68.85 | 88.71 |
| | | 4/4 | 67.36 | 87.76 |
| | | 1/4 | 64.36 | 85.65 |
| | Yang *et al.* [178] | 32/2 | 65.7 | 86.5 |
| | | 5/32 | 70.6 | 89.6 |
| | | 2/32 | 69.1 | 88.5 |
| | | 1/32 | 66.5 | 87.3 |
| | | 1/2 | 63.4 | 84.9 |
| | | 1/1 | 53.6 | 75.3 |
| | KDE-KM [131] | 4/32 | 61.82 | 83.89 |
| | LSQ [179] | 4/4 | 71.1 | 90 |
| | | 3/3 | 70.2 | 89.4 |
| | | 2/2 | 67.6 | 87.6 |
| | DSQ [147] | Mixed | 69.27 | - |
| | AdaRound [180] | 4/32 | 68.71 | - |
| | | 4/8 | 68.55 | - |
| | BitPruning [124] | 3.38/4.14 | 69.19 | - |
| | Bi-RealNet [25] | 1/1 | 56.4 | 79.5 |
| | RBNN [175] | 1/1 | 59.9 | 81.9 |
| | LNS [146] | 1/1 | 59.4 | 81.7 |
| | IR-Net [125] | 1/32 | 66.5 | 86.8 |
| | SLB [177] | 2/32 | 68.4 | 88.1 |
| | | 2/8 | 68.2 | 87.7 |
| | | 2/4 | 67.5 | 87.4 |
| | | 2/2 | 66.1 | 86.3 |
| | | 1/32 | 67.1 | 87.2 |
| | | 1/8 | 66.2 | 86.5 |
| | | 1/4 | 66 | 86.4 |
| | | 1/2 | 64.8 | 85.5 |
| | | 1/1 | 61.3 | 83.1 |
| | AdaRound [180] | 4/32 | 68.71 | - |
| | | 4/8 | 68.55 | - |
| | BRECQ [145] | 4/32 | 70.7 | - |
| | | 4/4 | 69.6 | - |
| | | 3/32 | 69.81 | - |
| | | 2/32 | 66.3 | - |
| | | 2/2 | 64.8 | - |

| Network | Method | Bit-width(W/A) | Top-1 (%) | Top-5 (%) |
|---|---|---|---|---|
| | DGRL [169] | 1/1 | 60.45 | - |
| | DMBQ [126] | 2/32 | 70.1 | 89.3 |
| | | 1/32 | 65.9 | 87.1 |
| | | 3/3 | 70 | 89.4 |
| | | 2/2 | 67.8 | 88.1 |
| | | 1/2 | 63.5 | 85.5 |
| | Liu et al. [159] | 4.38/4.38 | 70.59 | - |
| | | 3.55/3.55 | 70.12 | - |
| | | 2.72/2.72 | 69.84 | - |
| | HAWQ-V3 [150] | 8/8 | 71.56 | - |
| | | 4.8/4.8 | 70.22 | - |
| | | 4/4 | 68.45 | - |
| | BASQ [181] | 4/4 | 72.56 | - |
| | | 3/3 | 71.4 | - |
| | | 2/2 | 68.6 | - |
| | N2UQ [182] | 4/4 | 72.9 | 90.9 |
| | | 3/3 | 71.9 | 90.5 |
| | | 2/2 | 69.4 | 88.4 |
| | Bit-split [183] | 6/6 | 69.6 | - |
| | | 4/8 | 69.1 | - |
| | | 4/4 | 67.6 | - |
| | Oh et al. [184] | 4/32 | 69.46 | - |
| | | 3/32 | 68.76 | - |
| | | 2/32 | 65.62 | - |
| | Tang et al. [160] | 4/4 | 70.8 | - |
| | | 3/3 | 69.7 | - |
| | | 2.5/3 | 68.7 | - |
| | PD-Quant [185] | 4/4 | 69.23 | - |
| | | 4/2 | 58.17 | - |
| | | 2/4 | 65.17 | - |
| | | 2/2 | 53.14 | - |
| **ResNet-34 [8]** | Full-precision | 32/32 | 73.27 to 74.9 | 91.26 to 91.8 |
| | Venkatesh et al. [128] | 2/32 | 71.6 | 90.37 |
| | HWGQ [116] | 1/2 | 64.3 | 85.7 |
| | ABC [117] | 1/1 | 68.4 | 88.2 |
| | LQ-Nets [122] | 3/3 | 71.9 | 90.2 |
| | | 2/2 | 69.8 | 89.1 |
| | | 1/2 | 66.6 | 86.9 |
| | BCGD [137] | 4/8 | 72.18 | 90.73 |
| | | 4/4 | 70.81 | 90 |
| | | 1/4 | 68.43 | 88.29 |
| | LSQ [179] | 4/4 | 74.1 | 91.7 |
| | | 3/3 | 73.4 | 91.4 |
| | | 2/2 | 71.6 | 90.3 |
| | Bi-RealNet [25] | 1/1 | 62.2 | 83.9 |
| | RBNN [175] | 1/1 | 63.1 | 84.4 |
| | IR-Net [125] | 1/32 | 70.4 | 89.5 |
| | DMBQ [126] | 2/2 | 72.1 | 90.7 |
| | | 1/2 | 69.8 | 89.2 |

| Network | Method | Bit-width(W/A) | Top-1 (%) | Top-5 (%) |
|---|---|---|---|---|
| | N2UQ [182] | 4/4 | 76 | 92.8 |
| | | 3/3 | 75.2 | 92.3 |
| | | 2/2 | 73.3 | 91.2 |
| **ResNet-50 [8]** | Full-precision | 32/32 | 73.22 to 77.5 | 91.24 to 93.4 |
| | Venkatesh *et al.* [128] | 2/32 | 73.85 | 91.8 |
| | HWGQ [116] | 1/2 | 64.6 | 85.9 |
| | ABC [117] | 1/1 | 70.1 | 89.7 |
| | INQ [118] | 5/32 | 74.81 | 92.45 |
| | FGQ [119] | 2/8 | 70.76 | - |
| | | 2/4 | 68.38 | - |
| | Zuang *et al.* [24] | 4/4 | 75.7 | 92 |
| | | 2/2 | 70 | 87.5 |
| | PACT [107] | 32/4 | 75.9 | 92.9 |
| | | 5/5 | 76.7 | 93.3 |
| | | 4/4 | 76.5 | 93.2 |
| | | 3/3 | 75.3 | 92.6 |
| | | 2/4 | 74.5 | 91.9 |
| | | 2/2 | 72.2 | 90.5 |
| | | 1/2 | 67.8 | 87.9 |
| | LQ-Nets [122] | 4/32 | 76.4 | 93.1 |
| | | 4/4 | 75.1 | 92.4 |
| | | 3/3 | 74.2 | 91.6 |
| | | 2/32 | 75.1 | 92.3 |
| | | 2/2 | 71.5 | 90.3 |
| | | 1/2 | 68.7 | 88.4 |
| | Yang *et al.* [178] | 5/32 | 76.4 | 93.2 |
| | | 2/32 | 75.2 | 92.6 |
| | | 1/32 | 72.8 | 91.3 |
| | LSQ [179] | 8/8 | 76.8 | 93.4 |
| | | 4/4 | 76.7 | 93.2 |
| | | 3/3 | 75.8 | 92.7 |
| | | 2/2 | 73.7 | 91.5 |
| | HAWQ [154] | 2/4 | 75.48 | - |
| | HAWQ-V2 [155] | 2/4 | 75.76 | - |
| | SAT [186] | 4/32 | 76.4 | 93 |
| | | 3/32 | 76.3 | 93 |
| | | 2/32 | 75.3 | 92.4 |
| | | 4/4 | 76.9 | 93.3 |
| | | 3/3 | 76.6 | 93.1 |
| | | 2/2 | 74.1 | 91.7 |
| | AdaRound [180] | 4/32 | 75.23 | - |
| | | 4/8 | 75.01 | - |
| | Bi-RealNet [25] | 1/1 | 62.6 | 83.9 |
| | PWLQ [187] | 8/8 | 76.1 | - |
| | | 6/8 | 76.08 | - |
| | | 4/8 | 75.62 | - |
| | | 4/4 | 74.85 | - |
| | AdaRound [180] | 4/32 | 75.23 | - |
| | | 4/8 | 75.01 | - |

| Network | Method | Bit-width(W/A) | Top-1 (%) | Top-5 (%) |
| --- | --- | --- | --- | --- |
| | BRECQ [145] | 4/32 | 76.29 | - |
| | | 4/4 | 75.05 | - |
| | | 3/32 | 75.61 | - |
| | | 2/32 | 72.4 | - |
| | | 2/4 | 70.29 | - |
| | HAWQ-V3 [150] | 8/8 | 77.58 | - |
| | | 4.8/4.8 | 76.73 | - |
| | | 4/4 | 74.24 | - |
| | N2UQ [182] | 4/4 | 78 | 93.9 |
| | | 3/3 | 77.5 | 93.6 |
| | | 2/2 | 75.5 | 92.3 |
| | Bit-split [183] | 6/6 | 76.2 | - |
| | | 4/8 | 75.6 | - |
| | | 4/4 | 73.7 | - |
| | Oh et al. [184] | 4/32 | 75.75 | - |
| | | 3/32 | 75.14 | - |
| | | 2/32 | 72.27 | - |
| | Tang et al. [160] | 3/4 | 76.9 | - |
| | PD-Quant [185] | 4/4 | 75.16 | - |
| | | 4/2 | 64.18 | - |
| | | 2/4 | 70.77 | - |
| | | 2/2 | 57.16 | - |
| **ResNet-101 [8]** | Full-precision | 32/32 | 77.5 to 78.2 | 94.1 |
| | FGQ [119] | 2/8 | 73.85 | - |
| | | 2/4 | 70.69 | - |
| | LSQ [179] | 4/4 | 78.3 | 94 |
| | | 3/3 | 77.5 | 93.6 |
| | | 2/2 | 76.1 | 92.8 |
| | Bit-split [183] | 6/6 | 77.5 | - |
| | | 4/8 | 76.9 | - |
| | | 4/4 | 74.7 | - |
| **ResNet-152 [8]** | Full-precision | 32/32 | 76.5 to 78.9 | 93.2 to 94.3 |
| | Venkatesh *et al.* [128] | 2/32 | 76.64 | 93.2 |
| | LSQ [179] | 4/4 | 78.5 | 94.1 |
| | | 3/3 | 78.2 | 93.9 |
| | | 2/2 | 76.9 | 93.2 |
| | Bi-RealNet [25] | 1/1 | 64.5 | 85.5 |
| **MobileNetV1 [13]** | Full-precision | 32/32 | 70.9 to 72.4 | 89.9 to 90.2 |
| | SAT [186] | 4/32 | 72.1 | 90.2 |
| | | 3/32 | 70.7 | 89.5 |
| | | 2/32 | 66.3 | 86.8 |
| | | 8/8 | 72.6 | 90.7 |
| | | 6/6 | 72.3 | 90.4 |
| | | 5/5 | 71.9 | 90.3 |
| | | 4/4 | 71.3 | 89.9 |
| | Phan et al. [188] | 1/1 | 60.9 | 86.4 |
| | ReActNet [189] | 1/1 | 69.4 | - |
| | AdaBits [151] | 8/8 | 72.4 | - |
| | | 6/6 | 72.4 | - |

| Network | Method | Bit-width(W/A) | Top-1 (%) | Top-5 (%) |
|---|---|---|---|---|
| | | 5/5 | 72.1 | - |
| | | 4/4 | 71.1 | - |
| | BASQ [181] | 4/4 | 72.05 | - |
| | Tang et al. [160] | 4/32 | 72.6 | 90.83 |
| | | 4/4 | 71.84 | 90.38 |
| | | 3/32 | 71.57 | 90.3 |
| | | 3/3 | 69.48 | 89.11 |
| **MobileNetV2 [190]** | Full-precision | 32/32 | 71.72 to 72.62 | 90 to 90.5 |
| | DSQ [147] | 4/4 | 64.8 | - |
| | SAT [186] | 4/32 | 72.1 | 90.6 |
| | | 3/32 | 71.1 | 89.9 |
| | | 2/32 | 66.8 | 87.2 |
| | | 8/8 | 72.5 | 90.7 |
| | | 6/6 | 72.3 | 90.6 |
| | | 5/5 | 72 | 90.4 |
| | | 4/4 | 71.1 | 90 |
| | DJPQ [71] | Mixed | 69.3 | - |
| | AdaBits [151] | 8/8 | 72.6 | - |
| | | 6/6 | 72.4 | - |
| | | 5/5 | 72.1 | - |
| | | 4/4 | 70.8 | - |
| | AdaRound [180] | 4/32 | 69.78 | - |
| | | 4/8 | 69.25 | - |
| | BitPruning [124] | 4.15/4.57 | 70.09 | - |
| | BASQ [181] | 4/4 | 71.98 | - |
| | | 3/3 | 70.25 | - |
| | | 2/2 | 64.71 | - |
| | PD-Quant [185] | 4/4 | 68.19 | - |
| | | 2/4 | 55.17 | - |
| **SqueezNet [12]** | Full-precision | 32/32 | 69.38 | - |
| | HAWQ [154] | 3/8 | 68.02 | - |
| | HAWQ-V2 [155] | 3/8 | 68.38 | - |
| **DenseNet-121 [191]** | Full-precision | 32/32 | 75 | 92.3 |
| | LQ-Nets [122] | 2/2 | 69.6 | 89.1 |

It is inferred from Table 6 that some weight quantization methods, with 1-bit or 2-bit quantization, have achieved an accuracy close to the full-precision model when full-precision activations are used. When both weights and activations are quantized, a few methods with 4-bit or 3-bit precision have achieved higher accuracy than the full-precision model. For instance, the BASQ [181] and N2UQ [182] methods have achieved higher accuracy than the full-precision model for ResNet-18 with 4-bit precision. However, as the bit-width is further reduced to 2, and 1, the accuracy decreases significantly, especially in deeper networks. Despite the accuracy reduction, some methods, such as PACT [107], LQ-Nets [122], DMBQ [126], SLB [177], LSQ [179], BASQ [181], and N2UQ [182], have achieved accuracy levels close to that of the full-precision model. Figures 16, 17, and 18 present comparisons of top-1 accuracy for several methods with low-precision weights and activations for ResNet-18 on ImageNet, including XNOR-Net [22], Bi-RealNet [25], Balanced [79], PACT [107], HWGQ [116], LQ-Nets [122], DMBQ [126], BRECQ [145], LNS [146], RBNN [175], SLB [177], Yang et al. [178], LSQ [179], BASQ [181], N2UQ [182], and PD-Quant [185]. The accuracy of the full-precision model is taken from [182].

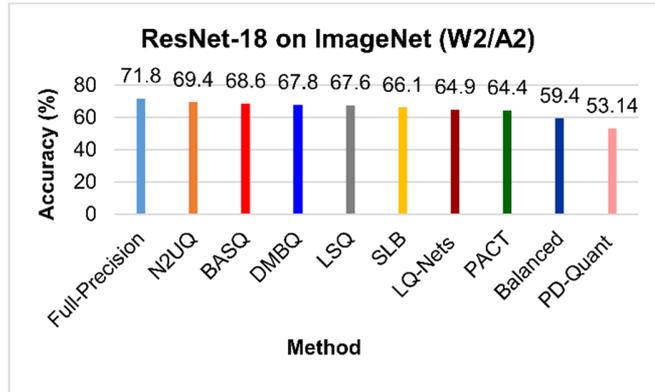

Figure 16: Top-1 accuracy of several methods in 2-bit weights and activations quantization for ResNet-18 on ImageNet

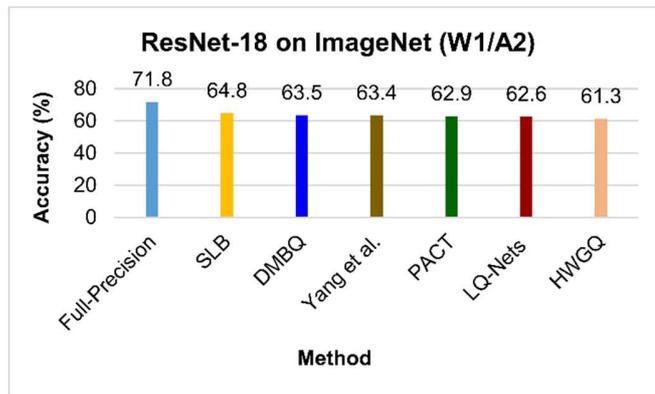

Figure 17: Top-1 accuracy of several methods in 1-bit weights and 2-bit activations quantization for ResNet-18 on ImageNet

In Figure 16, where weights and activations are quantized in 2-bit precision, it is observed that N2UQ [182], BASQ [181], DBMQ [126], and LSQ [179] have achieved accuracy close to the full-precision model. Figure 17 compares the models in 1-bit weights quantization and 2-bit activations quantization, while Figure 18 demonstrates the accuracy of some models in 1-bit weights and activations quantization. As Figure 18 shows, the accuracy considerably decreases in a binary model compared to the full-precision model. The best result is achieved by the SLB method [177], which is 8% lower than the accuracy of the full-precision model. ResNet-18 includes 18 layers, the accuracy gap in deeper Networks can be even more significant.

Regarding the reported accuracy in Tables 5 and 6, it is worth noting that quantization is conducted in different situations in various studies. For example, the first and last layers are quantized in some works, while some other works keep them in full precision. Therefore, for an accurate comparison between these works, it is crucial to refer to original papers and consider the details of their respective methodologies.

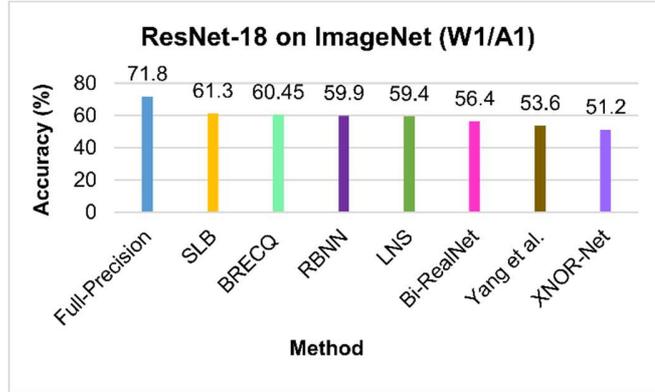

Figure 18: Top-1 accuracy of several methods in 1-bit weights and activations quantization for ResNet-18 on ImageNet

## 8 CONCLUSION AND FUTURE WORKS

In this paper, we surveyed the previous quantization works in the image classification task. The basic and advanced concepts of DCNN quantization were discussed, as well as the most important methods and approaches in this field, along with their advantages and challenges. Some previous works perform quantization on both weights and activations, offering a higher compression rate and employing lower-cost operations compared to approaches that are quantized only weights. However, quantization of activations is more challenging compared to weights, which is due to the wide range of activations, the use of a non-differentiable activation function, the estimation of activations during the backward pass, and the variation of activation values during inference.

The QAT and PTQ methods were studied, and it is concluded that the QAT methods generally achieve higher accuracy than the PTQ methods in the inference phase. Training a quantized DNN poses new challenges compared to a full-precision network since the units are discrete. It commonly requires additional iterations for convergence in contrast to training a full-precision network, and adaptive training strategies are required to build an accurate model. For instance, the adjustment of learning rate and regularization techniques can be different from the training in the full-precision network.

We discussed uniform and non-uniform quantization techniques and concluded that non-uniform quantization, especially the POT quantization approach, efficiently covers the distribution of full-precision values, which leads to enhanced accuracy. For decreasing quantization error, it is important to allocate quantization levels to informative regions. Using the scale factor helps in shifting the quantization levels to the most informative parts of data.

Some previous methods have successfully achieved high accuracy on large-scale datasets, such as ImageNet, when both weights and activations are quantized in low bit-width. However, quantization with a precision lower than 4 bits remains a challenging task, especially in deeper networks. During the training of a quantized network, STE is commonly used for calculating gradients in the backward pass. The noise resulting from gradient mismatch, due to inaccurate estimation, is amplified layer by layer from the end of the network to the initial layers. This amplification of noise is more considerable in deeper networks compared to shallow networks. In the training, this noise can have a negative impact on model convergence. Additionally, since the number of parameters increases with the depth of the neural network, the range of parameters in the deeper networks is wider than in shallow ones, and the quantization is more challenging. Accordingly, future works should focus on addressing the quantization of weights and activations in deeper networks with low bit-width, such as binary or ternary quantization.

In this paper, we discussed mixed-precision, which is currently an interesting approach in the quantization of the DNNs. The main challenge in mixed-precision quantization is the exponential time complexity in finding the optimum bit-width for each layer. It is desirable for future works to develop solutions that can determine the optimum mixed-precision with polynomial time complexity.